\documentclass[9pt,twocolumn,twoside]{osajnl}

\usepackage{times}
\usepackage{dblfloatfix}
\usepackage{amsmath}
\usepackage{amssymb}
\usepackage{multirow}
\usepackage{multirow}
\usepackage{mathtools}
\usepackage{units}
\usepackage{subfloat}
\usepackage{float}
\usepackage[]{units}
\usepackage{xcolor}
\usepackage{url}
\usepackage{bbm}
\usepackage{algorithm}
\usepackage{algpseudocode}
\usepackage{pbox}
\usepackage{placeins}
\usepackage{bm}
\usepackage{acronym}
\usepackage{enumitem}

\usepackage{booktabs}
\usepackage{graphicx}
\usepackage{caption}
\usepackage{newfloat}
\usepackage{times}
\usepackage{amsmath}
\usepackage{amssymb}
\usepackage{caption}
\usepackage{bm}
\usepackage{url}
\usepackage{hyperref}
\usepackage[capitalise]{cleveref}
\journal{ol} 
\usepackage{amsmath}
\usepackage{empheq}
\usepackage{multirow}
\usepackage{bm}
\usepackage{graphicx}
\usepackage{amssymb}
\newcommand\Set[2]{\{\,#1\mid#2\,\}}
\newcommand{\overbar}[1]{\mkern 1.5mu\overline{\mkern-1.5mu#1\mkern-1.5mu}\mkern 1.5mu}
\newcommand\scalemath[2]{\scalebox{#1}{\mbox{\ensuremath{\displaystyle #2}}}}
\setboolean{shortarticle}{false}

\title{Versatile Multi-Contact Planning and Control for Legged Loco-Manipulation}

\author[1,*]{Jean-Pierre Sleiman}
\author[1]{Farbod Farshidian}
\author[1]{Marco Hutter}

\affil[1]{Robotic Systems Lab, ETH Zurich, Zurich, Switzerland}
\affil[*]{Corresponding author: jsleiman@ethz.ch}

\begin{abstract}
Loco-manipulation planning skills are pivotal for expanding the utility of robots in everyday environments. These skills can be assessed based on a system's ability to coordinate complex holistic movements and multiple contact interactions when solving different tasks. However, existing approaches have been merely able to shape such behaviors with hand-crafted state machines, densely engineered rewards, or pre-recorded expert demonstrations. Here, we propose a minimally-guided framework that automatically discovers whole-body trajectories jointly with contact schedules for solving general loco-manipulation tasks in pre-modeled environments. The key insight is that multi-modal problems of this nature can be formulated and treated within the context of integrated Task and Motion Planning (TAMP). An effective bilevel search strategy is achieved by incorporating domain-specific rules and adequately combining the strengths of different planning techniques: trajectory optimization and informed graph search coupled with sampling-based planning. We showcase emergent behaviors for a quadrupedal mobile manipulator exploiting both prehensile and non-prehensile interactions to perform real-world tasks such as opening/closing heavy dishwashers and traversing spring-loaded doors. These behaviors are also deployed on the real system using a two-layer whole-body tracking controller.
\end{abstract}
\usepackage{newfloat}

\DeclareFloatingEnvironment{Movie}

\begin{document}

\maketitle

\section*{Introduction}

\noindent Mobile manipulators have been gaining considerable attention as we move towards integrating robotic systems into our unstructured and complex world. The reason is primarily rooted in their ability to unify the functionalities offered by fixed-base robotic manipulators and mobile platforms into a single system. A mobile base indefinitely extends a manipulator's workspace, whereas a manipulator promotes a mobile robot into an agent that can interact with its environment and actively modify it. Such a synergy enables these robots to cover a wide range of tasks akin to those tackled by a human -- nature's most versatile mobile manipulator.     

Numerous mobile manipulators have been developed in recent years, encompassing systems with diverse morphologies and varying degrees of autonomy. However, none has come close to matching human-level versatility in handling generic loco-manipulation problems. Loco-manipulation is a form of manipulation inherently involving a locomotion element. It is fundamentally a multi-contact planning and control problem where the robot should properly exploit and coordinate contacts with its surroundings to simultaneously manipulate itself (to move and maintain balance) and other objects. One of the most impressive displays of such skills was demonstrated by Boston Dynamics' quadrupedal robot SpotMini \cite{SpotWithArm} autonomously opening and navigating through a spring-loaded door. Although the company's work is unpublished, the complexity associated with planning and executing such a task is evident. Typically, a substantial amount of engineering effort goes into hand-crafting similar task plans in an elaborate state machine that composes a feasible sequence of connected sub-goals. The success of the full scheme essentially relies on the robot's ability to perform the proper whole-body motions and apply the necessary contact forces such that all sub-goals are fulfilled. Additionally, various lower-level objectives must be satisfied, such as maintaining balance and stability, being robust to external disturbances, respecting the system's physical limits, avoiding self-collisions, and avoiding collisions with the manipulated object and static obstacles. Developing a framework that can automatically and holistically resolve these problems remains an active research endeavor.                  

A vast literature on loco-manipulation planning exists wherein different kinds of platforms and strategies have been adopted for various applications. Conducting a general yet brief overview, we come across a broad collection of interesting examples: Aerial manipulators pushing doors \cite{AerialDoorOpening} or movable structures \cite{AerialObjectPushing}. Wheeled mobile manipulators wiping surfaces \cite{Mabi}, plastering walls \cite{ReconfigurableBase}, or opening doors and drawers \cite{WheeledDoorOpening,AdaptiveDoorOpening,RobustDoorOpening}. Humanoids moving large and heavy objects \cite{HumanoidPosturePlanningHeavyPushing}, stacking boxes \cite{TSR}, or manipulating articulated objects \cite{NAODoorOpening,HumanoidSentis}. Quadrupedal platforms with robotic arms performing dynamic throwing \cite{BostonDynamicsDynamicManipulation} and grasping \cite{SpotZimmerman} maneuvers, opening heavy spring-loaded doors \cite{AlmaDario,UnifiedMPC}, or solving a set of tasks within a kitchen testbed \cite{MayankPaper}; and many more \cite{Sentis,Ballbot,AerialManipulation,NonprehensileTransportation,CentauroObjectPushing}. 

When considering the evolution of these systems' planning architectures, one notices a shift from decentralized approaches towards more holistic solutions. The former decomposes the entire problem into a hierarchy of smaller puzzles that are easier to manage individually. This has also been a popular strategy for pure locomotion control of poly-articulated systems such as legged robots \cite{CapturePoint,DarioZMP}. But designing independent yet tightly-coupled sub-modules is a highly task-dependent and heuristics-based process that requires arduous manual tuning. In contrast, whole-body formulations for loco-manipulation implicitly account for the interactions among the different sub-systems, are more intuitive to tune, and result in naturally-coordinated motions. However, existing whole-body techniques have been primarily developed for the unimodal planning case, where switching of manipulation modes is either non-existent or is pre-defined by a skilled engineer \cite{SpotWithArm}. Therefore, a more sophisticated framework that can produce multi-modal behaviors would ultimately allow us to deal with a broader class of problems.
          


Recently, some of the most prominent approaches to robotic control design have been widely dominated by data-driven techniques. Relying on data eliminates the need for accurate analytical models, which can be very difficult to acquire for systems operating in complex and unpredictable settings.  For instance, given the intricacy of modeling contact phenomena, model-free reinforcement learning (RL) methods have gained notable traction in contact-rich applications such as legged locomotion \cite{Jemin,RL_bipedal,RL_Taka}, dexterous manipulation \cite{RL_Dexterous1,RL_Dexterous2,RL_Dexterous3}, and legged mobile manipulation \cite{DeepWBC,RL_Yuntao,ma2023learning}. However, one disadvantage of model-free RL is that it typically involves an inefficient trial-and-error process which leads to long training times before attaining satisfactory performance. So rather than relying on real-world experience, simulators are often employed to generate realistic training data efficiently \cite{RL_Nikita}. When combined with strategies that mitigate the sim-to-real gap \cite{RL_SimToRealSurvey}, this can enable reliable transfer to hardware. Besides being sample inefficient, RL algorithms perform poorly in the absence of a well-defined, dense reward signal, particularly when tasked with long-horizon planning in complex environments \cite{RL_Atari,RL_PRM}. Therefore, experts could spend months in laborious reward-shaping and hyperparameter-tuning tailored to a specific task. One common way of overcoming this issue has been by bootstrapping RL with a form of imitation learning. For example, training can be guided by recorded human demonstrations \cite{RL_Dexterous1,RL_Dexterous2}, animal motion capture clips \cite{IL_locomotion,ImitateAndRepurpose}, or by an RL-trained teacher with access to privileged information \cite{RL_Joonho,RL_Taka,RMA}. Nevertheless, generating expert demonstrations for every newly encountered task is time-consuming, and motion retargeting is often challenging, especially when the robot's morphology differs from that of the demonstrator.                



\begin{figure*}[htb!]
\captionsetup{format=plain}
    \centering
    \makebox[0pt]{\includegraphics[keepaspectratio, width = 
\textwidth]{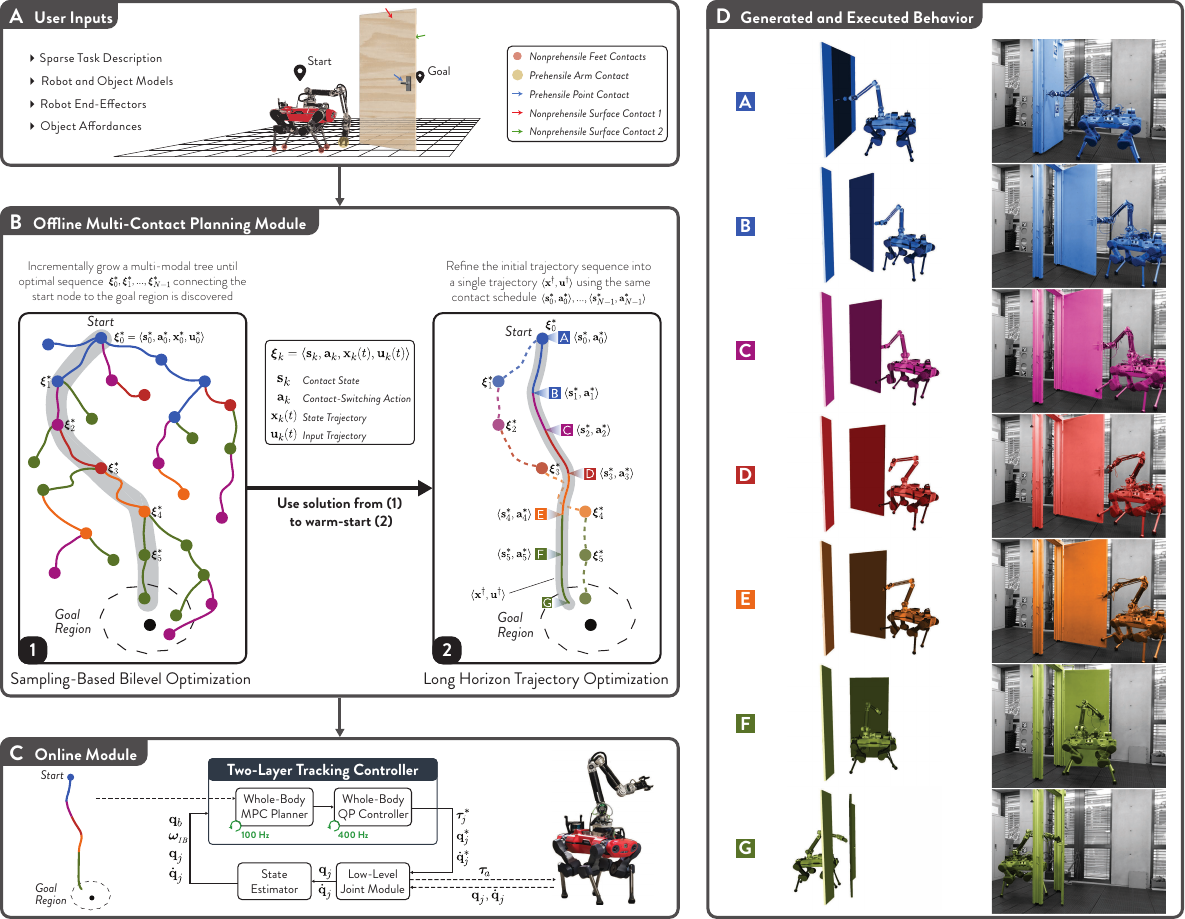}}
    \caption{\textbf{Planning and control architecture for multi-contact loco-manipulation.} (\textbf{A}) The user provides the offline planner with a basic high-level description of the scene: A sparse task specification given by a start state and a target goal region, dynamic models of the robot and the object, specification and classification of robot end-effectors and object affordances. (\textbf{B}) The central element in the offline planner is (\textbf{i}) a sampling-based bilevel optimization that gradually grows a tree of mode-invariant short-horizon trajectories seeking an optimal multi-modal sequence that reaches the goal region. Modes are determined by contact state and action pairs $\langle\bm s_k, \bm a_k\rangle$ that encode for each end-effector whether it is a closed, open, or switching manipulation contact, and the object contact it is interacting with. Each color corresponds to a specific contact state, and consecutive color variations indicate a contact switch. (\textbf{ii}) The second component in the planner is a long-horizon trajectory optimization that is warm-started with the bilevel optimization solution. It uses the same contact schedule and only polishes the continuous part. (\textbf{C}) To deploy multi-contact behaviors, the online module relies on an MPC-based whole-body controller that tracks the offline references. (\textbf{D}) With such a framework, complex behaviors that require exploiting multiple contact interactions, such as traversing a spring-loaded pull door, can be rapidly generated (i.e., often requires less than 1 minute on an ordinary laptop), and reliably executed with a real quadrupedal mobile manipulator.}
    \label{fig:figure1}
\end{figure*}

Data-driven control design has been a powerful tool for achieving robust policies. On the other hand, it does not offer a versatile framework that can systematically generalize well across diverse scenarios of a similar nature. With the aim of establishing such a framework for general loco-manipulation, we explore alternative approaches in the trajectory optimization literature. Discovering motion trajectories coupled with an optimal contact schedule is fundamentally a hybrid planning problem as it involves both continuous and discrete decision variables. In its most general form, multi-contact planning can be mathematically cast as a mixed-integer nonlinear program. Such programs are computationally intractable due to their inherent exponential complexity. Moreover, if provided with a poor initial guess, they typically end up in bad local minima arising from the problem's nonconvexity. To address these issues, some methods rely on approximations that convert the original problem into a mixed-integer convex program \cite{MIP_Locomotion,MIP_Manipulation}, offering global optimality guarantees upon convergence without requiring any form of warm-start. Alternatively, contact-implicit optimization (CIO) aims to resolve the combinatorial explosion of contact transitions by reformulating the problem as a continuous trajectory optimization. For instance, Mordatch \textit{et al.} \cite{TO_Mordatch,TO_MordatchManipulation} introduce contact-dependent costs weighted by real-valued auxiliary variables acting as continuous contact flags. These are then optimized jointly with the motion trajectories (and contact forces \cite{TO_MordatchManipulation}) within a single unconstrained nonlinear program (NLP). The approach of Mordatch \textit{et al.} \cite{TO_Mordatch} and other similar methods that eliminate discontinuities induced by contact dynamics \cite{TO_Neunert,TO_Tassa,TO_StochasticComplementarity} are essentially based on smooth relaxations of the contact complementarity system \cite{Anitescu}. These relaxations allow forces to be applied at a distance, resulting in gradients that guide the optimization routine toward contact-rich but physically-inconsistent behaviors. In contrast, Posa \textit{et al.} \cite{TO_Posa2} maintain the hybrid structure of contact by formulating the NLP as a mathematical program with complementarity constraints. They employ certain tricks that aid the solver's convergence despite the nonconvexity and nonsmoothness of the constraint manifold. Other notable works that implicitly and accurately handle contact phenomena pose the problem as a bilevel optimization: the outer level generates motion trajectories that are constrained by the resolution of the contact dynamics in the inner-level optimization \cite{TO_Carius,TO_Brian,BilevelDeformableContacts}. Nevertheless, in general, optimizations involving intricate physical models are highly sensitive to initialization and take time to converge, especially when planning through multiple contact switches. 


Complementary to the gradient-based perspective, we investigate the contributions of classical graph-search and sampling-based algorithms \cite{LaValle} in multi-contact planning. These can be seen as crucial assets when considering the fragility of optimization-driven methods in hybrid and non-convex scenarios. Early work by Dalibard \textit{et al.} \cite{SamplingBased_FullDoorOpening} addressed hybrid loco-manipulation tasks such as opening and navigating through doors by applying Rapidly exploring Random Trees (RRT) to find a sequence of robot-object motions intertwined with discrete transitions between successive constraints. A similar problem was later solved by searching through a graph of predefined motion primitives that provide feasible transitions between base configurations and discrete manipulation modes \cite{GraphBased_FullDoorOpening}. In both cases, the planners were tightly built around some form of task-specific knowledge. Hauser \textit{et al.} \cite{TAMP_MultiModal} proposed a more generic approach for multi-modal motion planning that builds a sparse tree of hybrid states by extending tree nodes towards randomly-sampled adjacent modes and robot configurations. More recently, Murooka \textit{et al.} \cite{GraphBased_LocoManipulationReachability} achieved general large-object loco-manipulation with humanoid robots by jointly planning sequences of footsteps and regrasps using a graph search with a transition model based on reachability maps. However, they rely on a Zero-Moment-Point (ZMP) balance criterion, which assumes coplanar contacts by neglecting the effects of manipulation forces. Looking at additional examples in the legged locomotion \cite{SamplingBased_EfficientAcyclic,SamplingBased_CDRM} or non-prehensile manipulation \cite{SamplingBased_WholeArmRearrangement,SamplingBased_ContactModeRRTDexterous} literature, it is evident that this family of multi-contact planners has been predominantly applied in the quasi-static domain. One way of extending them is via pre-specified dynamic motion primitives \cite{SamplingBased_PokeRRT,TAMP_MotionPrimitives1}. But such methods do not generalize well as they might require extra design effort to generate unique high-level primitives given a new task. These shortcomings have motivated synergizing search-based and sampling-based planners with trajectory optimization schemes \cite{GraphBased_LearnedCentroidal,TAMP_HybridPRM,GraphBased_TrajectoTree}. 

In the interest of unifying and formalizing concepts, we note that most of the above approaches can be classified as an instance of Task and Motion Planning (TAMP) problems \cite{TAMP_Garrett}.
In fact, the work by Hauser \textit{et al.} \cite{TAMP_MultiModal} can be interpreted as an extension of sampling-based motion planning to the TAMP domain. Analogously, Toussaint \textit{et al.} \cite{TAMP_LogicGeometric} proposed an optimization-based version, Logic Geometric Programming (LGP), that incorporates first-order logic to represent mode switches within a geometrically-constrained path optimization. This effectively amounts to a domain-specific extension of mixed-integer programs that can be handled more efficiently with a structure-exploiting solver. The LGP formalism was later adopted to tackle a wide range of sequential manipulation tasks \cite{TAMP_DifferentiableToolUse,TAMP_ObjectCentric,TAMP_DyadicManipulation,TAMP_Sydebo}, and also inspires the loco-manipulation framework presented in this paper.

We propose a versatile approach for generating high-fidelity multi-modal plans that solve long-horizon loco-manipulation tasks, under the assumption of pre-specified nominal object models and affordances. More specifically, our framework targets a class of problems involving general dynamic manipulation of movable or articulated objects with a mixture of prehensile and non-prehensile interaction modes. Given high-level descriptions of the robot and object, along with a task specification encoded through a sparse objective, our planner holistically discovers: how the robot should move, what forces it should exert, what limbs it should use, as well as when and where it should establish or break contact with the object. The framework can be readily adapted to different kinds of mobile manipulators. Nonetheless, for the sake of conciseness, we limit the discussion in this work to the case of a quadrupedal platform, ANYmal \cite{ANYmal,Anybotics}, equipped with a custom-built 6-DoF robotic arm (see \cref{fig:figure1}). Designing control architectures for such a high-dimensional (24-DoF) underactuated system exhibiting hybrid dynamics is a highly complex process.   

The diagram in \cref{fig:figure1} provides an overview of the proposed planning and control scheme. In our previous work \cite{UnifiedMPC}, we formulated a unified optimal control problem (OCP) that generates whole-body plans for combined locomotion and unimodal manipulation. Here, we build on this framework by extending it to accommodate various manipulation modes. This enables us to automatically compose a mixture of multi-modal plans without relying on task-dependent motion primitives. Manipulation modes are defined based on viable pairings between user-specified object affordances and robot end-effectors. As depicted in \cref{fig:figure1}A, we categorize affordances into prehensile/nonprehensile contact points or surfaces and end-effectors as prehensile/nonprehensile feet or arm contacts. The core insight is that such a classification enables us to encode feasible contact-switching actions via simple logic rules. This considerably reduces the branching factor in the discrete search and, accordingly, the computational complexity associated with traversing a connected graph of contact states. Therefore, inspired by the LGP formalism of TAMP problems \cite{TAMP_LogicGeometric}, we devise an offline planner in the form of a bilevel optimization \cite{TAMP_DyadicManipulation,TAMP_Sydebo} that employs a rule-based informed graph search at an outer level interleaved with an inner-level trajectory optimization for switched systems. Unlike most TAMP formulations, our sparse goal is not characterized in terms of a target symbolic state but rather as a terminal set of desired base-object poses. Furthermore, we aim to avoid the computational burden of searching through an extended graph that augments the contact state with a grid-based representation of our continuous state space \cite{TAMP_DyadicManipulation}. To this end, we adopt a sampling-based approach in generating the references for the inner-level problem, and incrementally grow a multi-modal tree of short-horizon trajectories by alternating between goal-directed and purely random extensions. This effectively covers the search space with a discrete set of reachable robot-object states coupled with their corresponding contact modes, as illustrated in \cref{fig:figure1}B (i). The introduced randomness also elevates the algorithm into a strategy capable of global exploration, an essential aspect for escaping bad local minima arising from the nonconvexity of geometric constraints.

After converging to a connected sequence of optimal trajectories and contact switches leading to the goal, we apply a post-processing step to enhance the solution's overall quality. As shown in \cref{fig:figure1}B (ii), this entails using the discovered sequence as an initial guess to warm-start a single long-horizon optimization over a fixed contact schedule. The resulting output is a smoother and lower-cost solution with increased feasibility, yielding a high-fidelity plan that can be reliably executed on the real system.  

Finally, to ensure a robust deployment of complex behaviors on such robotic systems, a common approach has been to decompose the problem into an offline planning phase that handles the heavy computations and a reactive control module that tracks the generated plans at high-update rates \cite{ROLOMA,TO_Winkler}. However, when dealing with long-horizon dynamic maneuvers involving multiple discontinuous switches, the accumulated effects of unmodeled disturbances might render the plan unstabilizable with a purely reactive controller. Hence, we mitigate these issues with a two-layer tracking scheme \cite{TO_Bjelonic,TO_AtlasBostonDynamics} consisting of a short-horizon Model Predictive Control (MPC) layer on top of a whole-body controller. The full architecture of the online module is depicted in \cref{fig:figure1}C. With such a structure, we attain a task-agnostic transfer of multi-contact plans with minimal tuning effort, thereby adequately bridging the gap between offline behavior generation and online execution. 

\begin{figure}[t]
    \centering
  \makebox[0pt]{\href{https://www.youtube.com/watch?v=rAP7M4BL9sQ&t=163s}{\includegraphics[width=\linewidth, keepaspectratio]{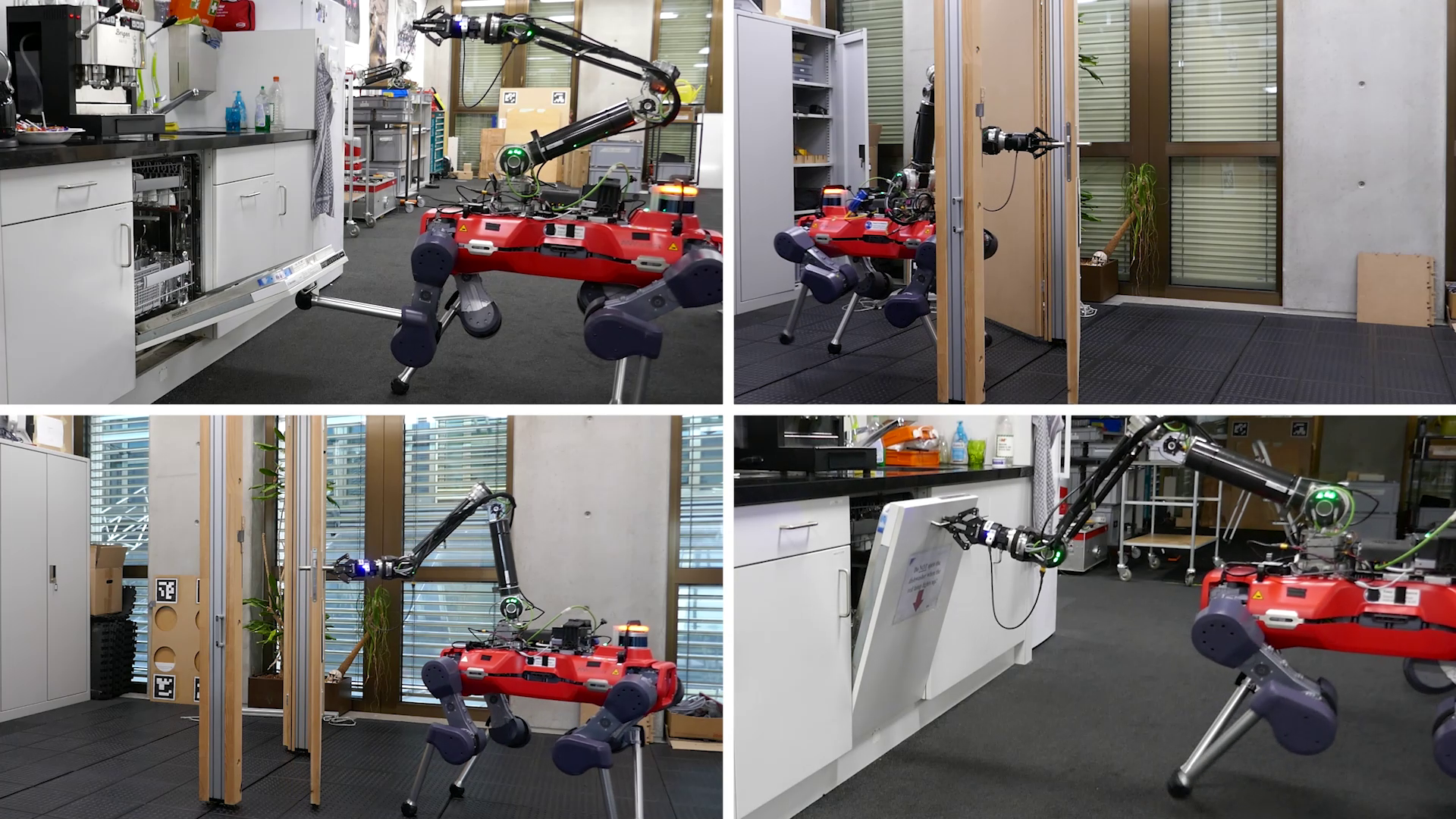}}}
\captionsetup{type=Movie}
\caption{\label{fig:plate} \textbf{Summary of the results and methods.}}
\end{figure}
     
We demonstrate the effectiveness of our framework in its ability to rapidly discover holistic solutions for a diverse set of tasks characterized by different objects, environments, or goal specifications. We verify the physical consistency of the offline behaviors by testing some of them on hardware and showing that, under the assumption of a reasonable object model, they can be accurately tracked by relying solely on proprioceptive feedback. One example, shown in \cref{fig:figure1}D, is that of a door-traversal sequence including multiple interaction modes.
\section*{RESULTS}

\noindent Movie 1 summarizes the methodology and results of the presented work. The following subsections describe the results in detail. 

\subsection*{Overview of loco-manipulation experiments}

\begin{figure*}[ht!]
\captionsetup{format=plain}
\includegraphics[keepaspectratio, width = 
\textwidth]{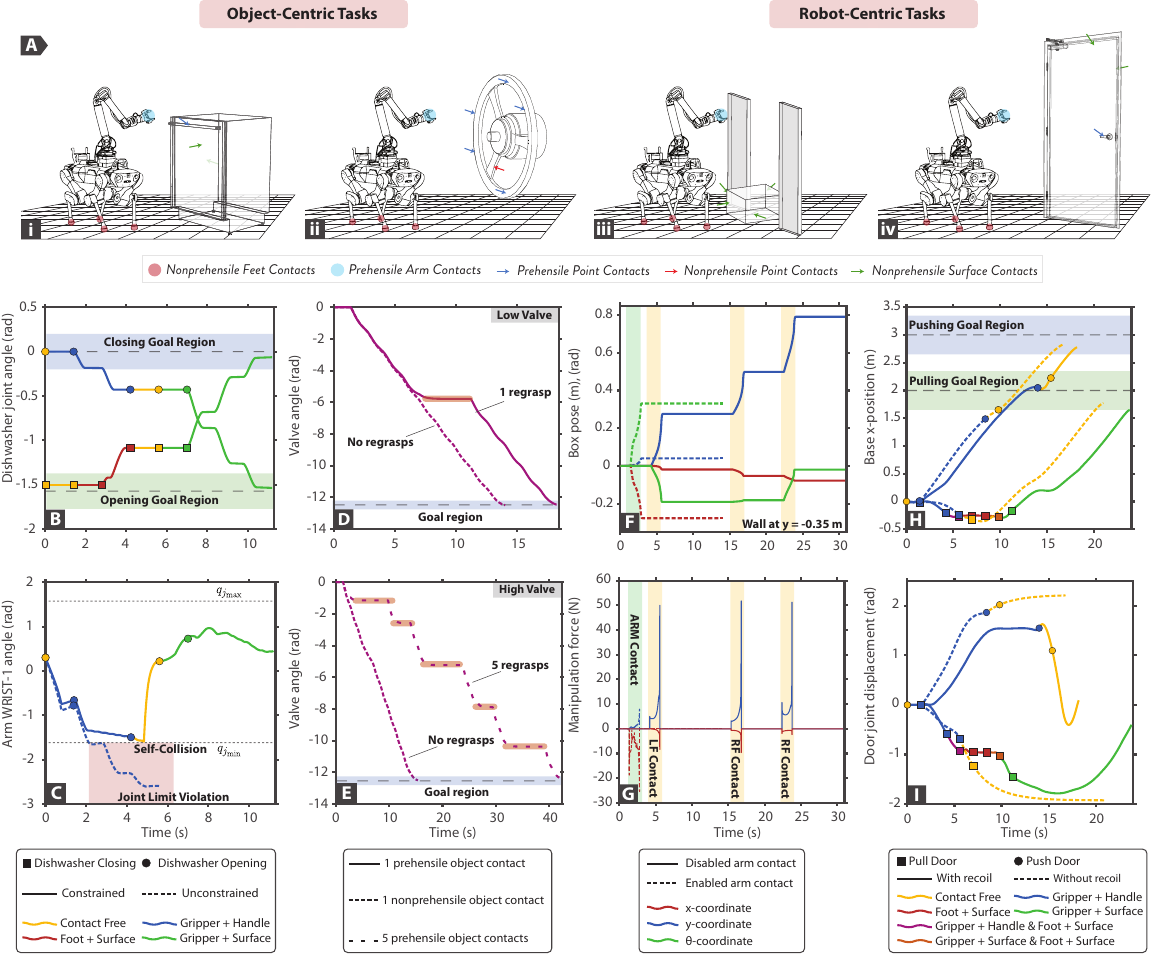}
\caption{\textbf{Validation of loco-manipulation planner.} (\textbf{A}) Illustration of two object-centric and two robot-centric tasks: i) dishwasher manipulation (\href{https://youtu.be/hgIanclikFI}{Movie S1}), ii) valve manipulation (\href{https://youtu.be/2YhAm4o7w3c}{Movie S2}), iii) traversal of an obstructing movable obstacle (\href{https://youtu.be/haS3y2uZDH0}{Movie S3}), iv) traversal of a large articulated object  (\href{https://youtu.be/-vLXUrBMbMI}{Movie S4}). Diverse behaviors emerge highlighting the planner's adaptability to variations in each task. \textbf{Task 1:} (\textbf{B}) Different start states and goal regions (opening vs. closing) (\textbf{C}) Opening with different joint limits and self-collision constraints for the robot. \textbf{Task 2:} (\textbf{D}) and (\textbf{E}) Different numbers, types and locations of object contacts, for two varying valve heights (low vs. high). \textbf{Task 3:} (\textbf{F}) and (\textbf{G}) Different robot end-effector contacts enabled for manipulation (enabled arm contact vs. blocked arm contact). \textbf{Task 4:} (\textbf{H}) and (\textbf{I}) Different joint limits (push vs. pull) and dynamics (recoil vs. no recoil) for the object.}
\label{fig:figure2}
\end{figure*}

\noindent Examining a wide variety of real-world loco-manipulation scenarios, we notice that they can be broadly categorized based on their high-level goal descriptions into two main groups: object-centric and robot-centric tasks. The former encompasses tasks centered around manipulating an object to alter its state towards a target configuration, such as opening a fridge, closing a dishwasher/oven, or turning a valve. On the other hand, a robot-centric perspective in the context of mobile manipulation entails a goal specification centered around the robot's base. This often requires the agent to interact with its environment as it navigates from one state to another. Examples of this class include traversing to the other side of a closed door or navigating among movable obstacles. Here, we opted for two representative tasks per category, as illustrated in \cref{fig:figure2}A. 

Through a series of simulation experiments, we establish the versatility of our planning framework by showcasing its ability to efficiently discover complex long-horizon behaviors with minimal manual guidance: All experiments assumed unknown task durations, sparse goal specifications, a uniform cost function, and no prior contact schedules or motion plans to guide the solver. Our results primarily illustrate how new behaviors automatically emerge from subtle variations in the task formulation -- such as changes in the robot's joint limits when opening a dishwasher, the object affordances when turning a valve, the available end-effectors when navigating across a movable obstacle, and the object dynamics when traversing a door.

To confirm the dynamic feasibility of the offline-generated plans, we conducted a set of hardware experiments for Tasks 1 and 4 of \cref{fig:figure2}A using our MPC-based tracking controller. Furthermore, we demonstrate the reliability of these plans, in terms of task-attainment, by successfully executing multiple independently-computed behaviors that solve the same task, but with variations in the problem parameters (such as the robot's desired gait schedule, and the object's dynamic model), or in the resulting manipulation schedule.

In the following section, we first analyze the emergent behaviors for all four tasks, and then present hardware results on our quadrupedal mobile manipulator. All discussed solutions correspond to the first goal-attaining sequence found by the planner.          

\subsection*{Behavior generation for object-centric tasks} 

\noindent In the first object-centric scenario, the robot manipulates a heavy dishwasher door that exhibits high stiction at the joint. As illustrated in \cref{fig:figure2}A (i), three potential object contacts are assumed: the front surface, the handle attached to it, and the back surface. All generated multi-modal solutions in \cref{fig:figure2} (B and C) are presented in the form of continuous trajectories coupled with a color-based encoding of the underlying manipulation modes. We started with the dishwasher fully closed and provided a goal region centered at -90 degrees. As shown in \cref{fig:figure2}B, the resulting behavior consists of the arm initially grasping the handle, slightly opening the dishwasher, then switching to the back surface-contact to complete the opening motion. The need for the intermediate contact switch can be explained by referring to \cref{fig:figure2}C, where its emergence is visibly justified by the arm's operational limits. In fact, by deactivating joint limits and self-collision constraints in our formulation, we observe that the contact switch is no longer necessary, as the dishwasher can be opened with a continuous gripper-handle interaction.

Next, we simply swapped the start and goal configurations of the dishwasher. Consequently, various closing maneuvers with diverse contact schedules are discovered, one of which is depicted in \cref{fig:figure2}B. The solution is a fully nonprehensile interaction sequence that exploits different end-effector contacts: The robot first uses one of its feet to lift the dishwasher, making it then easier for the arm to establish a nonprehensile contact with the same surface and thereby complete the closing task. 

The second object-centric task involves manipulating a large valve wheel with dynamics subject to static friction. A valve target of two full turns was assigned; however, we introduced various scenarios in which different object affordances (namely different numbers, types, and locations of contacts) were chosen to highlight their effects on the resulting manipulation behavior. First, we considered the case of a single nonprehensile point contact located at one of the spokes of the valve. \cref{fig:figure2}D indicates that a feasible plan consists of a continuous valve rotation by relying on a constant interaction. In contrast, replacing this contact with a prehensile one at the wheel's outer rim makes a regrasp necessary due to the additional constraint imposed on the gripper's orientation with respect to the graspable point which causes a violation in the limits of the arm's last joint. 

Next, we constructed a setting wherein the valve's height was adequately increased such that the inner nonprehensile point was still kinematically reachable for all valve angles, but the outer graspable contact was not. Therefore, in the prehensile interaction case, one object contact was no longer sufficient to fully solve the task. We resolved this by representing the valve affordance more accurately with multiple (five) grasping locations along the outer rim. As shown in \cref{fig:figure2}E, a viable turning sequence consists of several regrasps which occur whenever the arm gets close to its kinematic singularity throughout the manipulation period.   

\subsection*{Behavior generation for robot-centric tasks}    
\noindent In both robot-centric tasks, the base goal location is only attainable by altering the environment, which is composed of obstructive manipulable objects and other static obstacles. We did not specify any target regions for the object; hence, it can end up in an arbitrary terminal state.

The first scenario involves a movable obstacle modeled as a 2-kg box with three degrees of freedom: one rotational and two translational, along with four contact surfaces. In order to adequately capture frictional effects in all directions, we approximated the box-ground patch interaction with four point contacts at the bottom vertices of the cuboid. If the box is not large enough to block the robot's path, the planner finds the trivial locomotion-only solution of navigating around the object without having to manipulate it. On the other hand, in the case of an obstructing box, one of the resulting feasible plans employs the arm's end-effector in a single nonprehensile interaction phase that pushes the box sideways (y-direction). The displacement is large enough for the robot to clear its path while ensuring the object does not collide with the wall. 

Within the same setting, we considered another example wherein the arm cannot be used for manipulation, such as when the gripper is already carrying a load. Consequently, an alternative solution that exploits the robot's feet to displace the obstacle with a series of forward pushes (x-direction) emerges. These behaviors are captured in \cref{fig:figure2} (F and G), which show the object's motion and the corresponding manipulation forces exerted by the robot. The presented plots also highlight the dynamic feasibility of the generated trajectories: unilateral pushing forces, force-motion consistency, and stick-slip motion dynamics caused by static friction.   

In another scenario, the robot is required to traverse a large articulated object with nonlinear recoil dynamics, namely a spring-loaded door. We produced variations of the same task through basic adjustments in the door's workspace limits, articulation, and dynamic parameters. Affordances were specified as one graspable point (the door handle) and all pushable surface contacts. 

\newcommand\fillin[1][3cm]{\makebox[#1]{\dotfill}}
\begin{table*}[htbp]
\centering
\begin{tabular}{l|cccccc}
\textbf{Loco-Manipulation Scenario}           & \begin{tabular}[c]{@{}c@{}}\textbf{Behavior Minimum}\\ \textbf{Duration (s)}\end{tabular} & \multicolumn{4}{c}{\begin{tabular}[c]{@{}c@{}}\textbf{Planner Timing Statistics \textbf{(s)}}\\ {[}\textbf{Mean~~~~Min~~~~Max~~~~Std}{]} \end{tabular}} & \begin{tabular}[c]{@{}c@{}}\textbf{Tree-Extension Mean}\\ \textbf{Solve Time (ms)}\end{tabular} \\ \hline
\rule{0pt}{3ex}
Heavy Dishwasher Opening               & 9.8       & ~~~ 9.6      & 4.6   & 25.9  & \hspace{-0.3cm}9.2 & 226  \\
\rule{0pt}{3ex}
Heavy Dishwasher Closing               & 9.8     & ~~~ 13.6     & 7.5   & 25.8  &\hspace{-0.3cm}7.2  & 277   \\
\rule{0pt}{3ex}
Valve + 1 Non-Prehensile Contact   & 14     & ~~~ \textcolor{blue}{6.5}      & 5.7   & 7.2   & \hspace{-0.3cm}0.7   & 296   \\
\rule{0pt}{3ex}
Valve + 5 Prehensile Contacts     & \textcolor{blue}{42}        & ~~~ 19.4     & 16.5  & 21.6  &\hspace{-0.3cm} 2.2  &233  \\
\rule{0pt}{3ex}
Movable Obstacle + Arm Contact         & 14      & ~~~ 15.2     & 13.1  & 19.9  & \hspace{-0.3cm}2.7  & 349 \\
\rule{0pt}{3ex}
Movable Obstacle + Feet Contacts       & 30.8      & ~~~ 43.2     & 38.0  & 49.9  & \hspace{-0.3cm}4.7  & 337 \\
\rule{0pt}{3ex}
Push Door with Recoil                  & 15.4    & ~~~ 13.9     & 5.7   & 22.0  & \hspace{-0.3cm}7.1  & 256   \\
\rule{0pt}{3ex}
Pull Door with Recoil                  & 23.8     & ~~~ \textcolor{blue}{43.7}    & 33.8  & 53.9  & \hspace{-0.3cm}7.5 & 275  \\
\rule{0pt}{3ex}
Sliding Door with Recoil               & 26.6     & ~~~ 40.3     & 29.3  & 51.5  & \hspace{-0.3cm}8.0  & 284  \\
\rule{0pt}{3ex}
Push Door without Recoil               & 15.4      & ~~~ 8.1      & 4.8   & 11.9  & \hspace{-0.3cm}3.4 & 252  
\end{tabular}
\captionsetup{format=plain}
\caption{\textbf{Planner evaluation over ten scenarios.} Planner timing statistics were based on five test runs per scenario. Tree extensions correspond to solving a mode-invariant OCP with a short horizon of 1.4 seconds. The minimum and maximum average solve times, as well as the longest behavior duration are all highlighted in blue.}
\label{table1}
\end{table*}

Starting with the case of a spring-loaded revolving door that can be pushed open, the simplest solution to this task comprises: Establishing contact with the handle, maintaining this contact while pushing, and breaking the contact while ensuring the door does not collide with the robot upon recoil. By imposing an extra joint limit on the door ${(q_o \leq 0)}$), transforming it into one that can only be pulled open, a more complex multi-modal sequence arises; one of the feet is required to hold the door as the arm switches from the handle contact to the surface contact in the middle of the opening maneuver. Otherwise, it cannot fully pass through without violating the inherent geometric constraints. A similar schedule is also observed for the sliding-door case. The discovered behaviors are presented in \cref{fig:figure2} (H and I) in the form of base-object position trajectories with a color-based encoding of the underlying modes. 

Finally, by removing the door's recoil, the planner converges to the same straightforward sequence that solves all three examples. Indeed as shown in \cref{fig:figure2}I, the door can be easily opened and traversed using a single handle-contact interaction that is released at an early stage of the task duration. 

\subsection*{Planner evaluation}
\noindent We assessed the solver's performance considering ten of the discussed scenarios with five independent runs each. The related results mainly consist of the timing statistics relative to the corresponding behavior duration and are reported in \cref{table1}. The majority of the planner's computation is dedicated to the sampling-based bilevel optimization stage: for instance, in the pull-door scenario, around $82 \%$ of the computational time is spent in this stage, while the rest corresponds to the long-horizon trajectory optimization. Furthermore, we observe that had the proper manipulation schedule been pre-defined for this task, searching over the continuous trajectory segments would have only required around $13 \%$ of the bilevel search's total time. It is also worth noting that in contrast to our sampling-based method, a deterministic approach that accurately solves the bilevel problem through a grid-based representation of the continuous state space would have resulted in less variability in the computational times and higher-quality solutions that do not require any post-processing; however, it would have been less computationally efficient overall. 

All trajectories were computed on an ordinary laptop (Intel Core i7-10750H, 2.6 GHz, Hexa-core) in less than 1 minute for an average computational time of 279 ms per tree extension. Considering the planner's average solve times, normalized by the behavior durations, we find that the solver performs best for the valve-turning task and takes the most time to converge for the pull-door example, with a ratio of 0.46 and 1.84, respectively. 

\begin{figure}[ht!]
\captionsetup{format=plain}
\includegraphics[keepaspectratio, width = 
\linewidth]{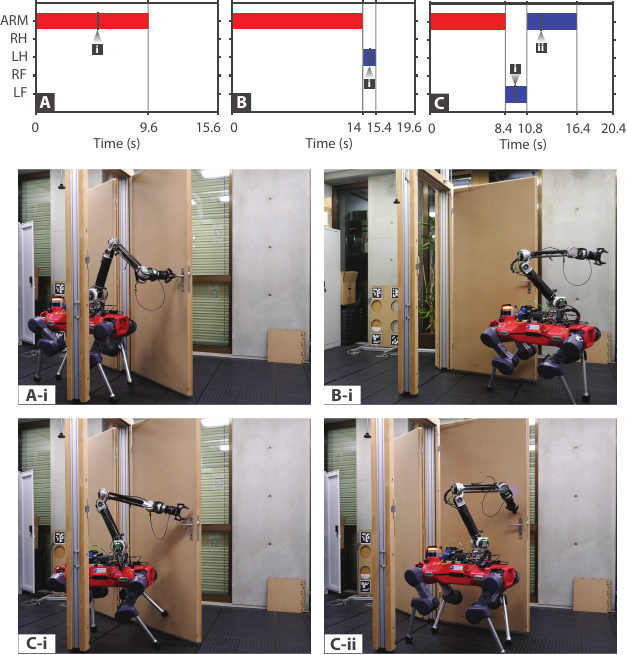}
\caption{\textbf{Hardware experiments -- Feasibility verification for push door with recoil behaviors.} (\textbf{A} to \textbf{C}) Different manipulation schedules discovered and executed to solve the same task. Red: prehensile interaction with handle, Green: nonprehensile interaction with surface. ARM: 6-DoF robotic arm, LF: left front leg, RF: right front leg, LH: left hind leg, RH: right hind leg. The numbers on the schedules indicate certain instants captured in the corresponding snapshots.}
\label{fig:figure3}
\end{figure}

We observe that the intricacy of the problem's dynamics doesn't substantially impact the planner's computational demands. This is indeed reflected in the tree-extension solve times which are mostly similar for all scenarios. Moreover, as a consequence of the adopted pruning rules, the planner's efficiency is not necessarily hindered by the number of total discrete modes, as demonstrated in the valve-manipulation scenario. Instead, we notice that the computational times are strongly correlated with the level of geometric complexity in the problem (namely, the nonconvexity of collision constraints) and the degree of sparsity in the goal specification. For instance, this is evident in the planner's performance when applied to the door traversal task. In contrast to the push-door scenario, traversing a pull-door requires the robot to initially move away from the goal and to also switch the arm contact across the object while avoiding collisions. Such behaviors underscore the need for an exploratory aspect in the planner.    


\begin{figure}[ht!]
\captionsetup{format=plain}
\includegraphics[keepaspectratio, width = 
\linewidth]{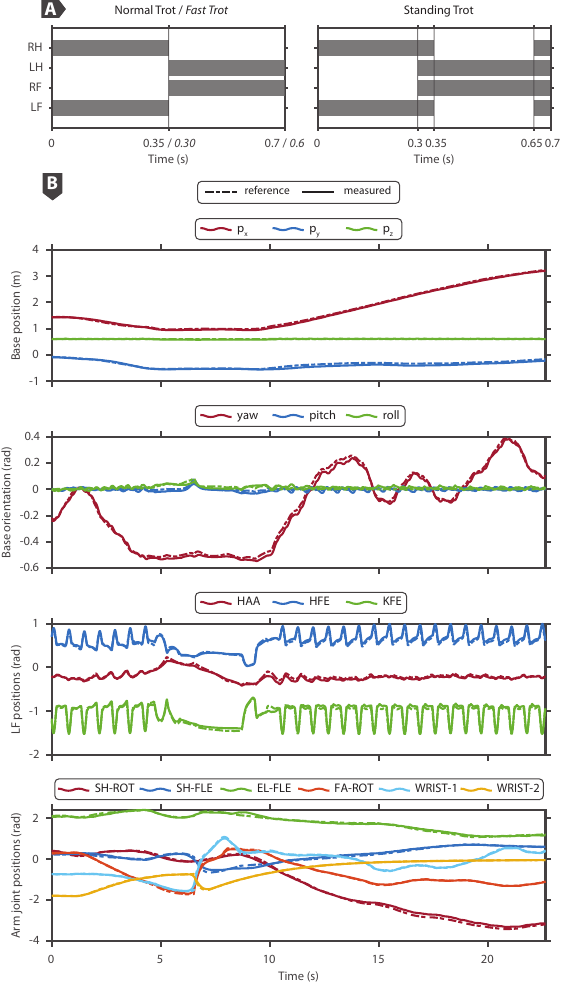}
\caption{\textbf{Hardware experiments -- Feasibility verification for pull door with recoil behaviors.} (\textbf{A}) The different gait schedules adopted in this scenario. Left: normal trot and fast trot; Right: standing trot. LF: left front leg, RF: right front leg, LH: left hind leg, RH: right hind leg. (\textbf{B}) Motion tracking accuracy of the base pose and manipulating limbs.}
\label{fig:figure4}
\end{figure}

\subsection*{Behavior execution}

\noindent Here, we focus on validating the physical consistency of the multi-modal behaviors by deploying them on hardware. In these experiments, the offline-generated trajectories were tracked by the real system using Module C, depicted in \cref{fig:figure1}. The two-layer tracking control and state estimation \cite{Bloesch} both run on ANYmal's main onboard PC (Intel i7-8850H, 2.6 GHz, Hexa-core). With a ${\text{1-second}}$ prediction horizon and a discretization time step of ${\text{0.015 s}}$, the MPC loop is executed at ${\text{100 Hz}}$ asynchronously to the whole-body controller and the state estimator, which are updated at a rate of ${\text{400 Hz}}$. The same hyperparameters were used in the MPC-WBC tracking module during all scenarios.

The experimental procedure simply consisted of precomputing the plans given a reasonable object model and storing them in a motion library on the robot's onboard PC. These trajectories were then mapped to the online setting by establishing a correspondence between a key reference frame in the real environment and its counterpart in the offline case. An example of such a frame is the end-effector as it initially grasps the door handle.

We selected the behaviors of highest complexity from one robot-centric and one object-centric task commonly encountered in real-world settings: traversing a spring-loaded door and manipulating a heavy dishwasher exhibiting high stiction. First, we tested the push-door case, which can be solved with a straightforward sequence described in the previous section. However, due to the sampling-based nature of our planner, we obtain various pushing maneuvers and manipulation schedules that can solve the task given the same goal specification. Three of the discovered mode sequences are depicted in \cref{fig:figure3}, two of which involve the use of one of the feet to hold the door as the arm either switches to a new object contact or retracts to its nominal configuration. 

Next, we showcase the more complex behavior of opening and passing through a pull door. So far, this has only been presented in \cite{SpotWithArm}, where laborious handcrafting appears to have been involved in the design process. The conducted experiments readily applied this sequence in two settings containing doors with different kinematic and dynamic parameters (different hinge side, width, inertia, and recoil) by solely changing the door's kinematic parameters in the planner formulation. Moreover, we note that varying multi-contact plans with the same manipulation sequence were generated by diversifying the gait schedules used when solving the task. These were all accurately tracked by the robot, where the adopted dynamic gaits included a fast trot, a normal trot, and a standing trot, as depicted in \cref{fig:figure4}A. 
The motion tracking accuracy of the robot's base and manipulating limbs is presented in \cref{fig:figure4}B for one of the pull-door solutions. The results indeed demonstrate that the online execution highly matches the offline references and that the transfer of precomputed trajectories to the physical system is agnostic to variations in the task formulation parameters.   

\begin{figure*}[ht!]
\captionsetup{format=plain}
\includegraphics[keepaspectratio, width = 
\textwidth]{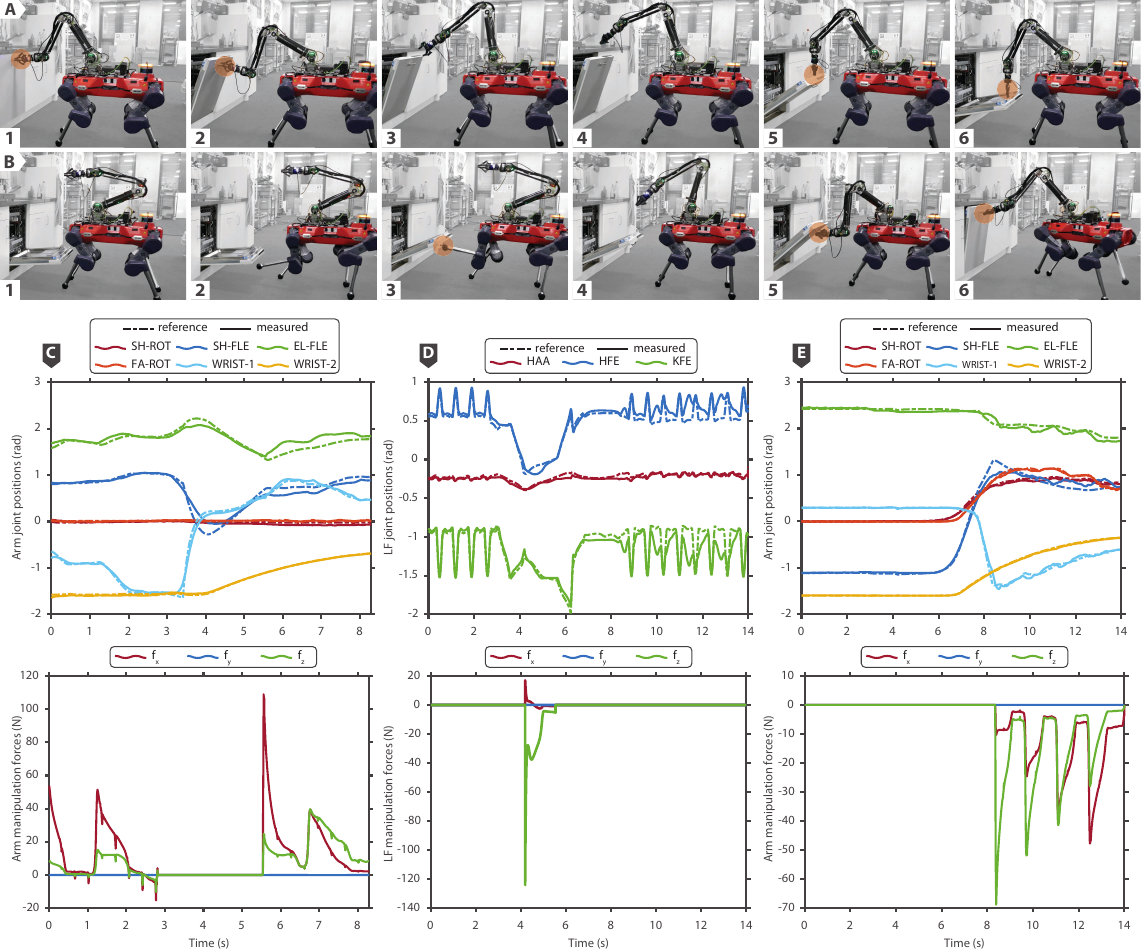}
\caption{\textbf{Hardware experiments -- Feasibility verification for dishwasher manipulation behaviors.} (\textbf{A}) and (\textbf{B}) Image sequence of executed multi-contact behaviors for manipulating a heavy dishwasher with high stiction highlighting different interaction phases: opening and closing maneuvers, respectively. (\textbf{C}) Motion tracking accuracy and exerted manipulation forces corresponding to the arm during the opening scenario. (\textbf{D}) and (\textbf{E}) Motion tracking accuracy and exerted manipulation forces corresponding to the arm and left front leg, respectively, during the closing scenario.}
\label{fig:figure5}
\end{figure*}

We further verify the framework's physical fidelity by manipulating a real dishwasher while exploiting multiple contact interactions, as shown in \cref{fig:figure5} (A and B). The tracking accuracy of the manipulating limbs for the opening and closing case is shown in the joint positions plots of \cref{fig:figure5}C and \cref{fig:figure5} (D and E), respectively. In such a scenario, the robot was operating close to its joint limits and experienced large manipulation forces of up to ${\text{125 N}}$ due to the high friction in the dishwasher's joint (see force plots of \cref{fig:figure5} (C to E)). 

The trajectories discussed in this section were successfully tested on the robot from the first trial without requiring any specific hand-tuning. Moreover, we note that each behavior was reliably executed in several test runs without any failures.    

\section*{Discussion} 

\noindent The framework presented in this paper achieved a rapid generation of holistic multi-modal behaviors along with their reliable execution in challenging scenarios on a high-dimensional legged system. Robot-centric and object-centric tasks consisting of basic high-level specifications, sparse objectives, and unknown time horizons were solved within a minute for task durations that reach up to 50 seconds. In the general context of contact-rich planning, several approaches exist in the locomotion/manipulation literature that, in contrast to ours, mostly lead to physically-inconsistent maneuvers, require long computational times, or are highly task-specific and hence involve a cumbersome design process. Apart from the introduction of domain-specific guiding rules, the success of the presented approach can be attributed to its exploitation of different planning techniques whose core strengths complement one another: Optimization-based planning for computing locally-optimal solutions that satisfy dynamics and path constraints, informed graph search for a fast and optimal combinatorial search, and sampling-based planning for an efficient global exploration. 

On the other hand, we see some limitations that we aim to address in future work. These limitations are primarily connected to the task-execution phase. First, the present capabilities of online replanning offered by the MPC-based controller are limited to solely adapting the continuous elements of the offline multi-modal references. Extending our framework to allow for an online adjustment of manipulation schedules necessitates making the bilevel optimization routine real-time capable. This could be achieved by speeding up the search through a proper parallelization of multiple tree extensions. 

Second, tracking behaviors generated on the basis of a nominal world model is only viable under the assumption of a reasonably accurate description; however, certain types of modeling mismatches can be tolerated more than others. For instance, during our hardware tests, we had to ensure that object articulations and geometries were captured with relatively high certainty; whereas we only provided a simple approximation of the dynamics that roughly represents the overall system response. Therefore, none of the real-world objects actually required an accurate system identification process when defining their corresponding equations of motion in the offline phase. In fact, the trajectories computed for the two distinct doors shown in \href{https://youtu.be/2H0op0mksQo}{Movie S6} rely on the same nominal model in the planner except for changes in the kinematic parameters (the door width and hinge side), despite them also having different masses and recoils. The reason is that if quasi-static effects are roughly compensated for, the unmodeled residual dynamics can be mitigated with high-gain feedback control. This inherent robustness is further demonstrated in \href{https://youtu.be/2H0op0mksQo}{Movie S6} through the controller's disturbance-rejection capabilities. In contrast, geometric and kinematic mismatches cannot be handled with our current approach and would potentially lead to dangerous collisions, or large internal forces caused by wrongly controlling position in motion-constrained directions.   

Nonetheless, we see the proposed framework as a stepping stone toward developing a fully autonomous loco-manipulation pipeline. For instance, robustness to large unforeseen disturbances and unmodeled effects can be greatly improved by complementing our planner with data-driven techniques. Specifically, reinforcement learning has been shown to produce policies with strong robustness properties when tailored to a specific objective \cite{RL_Joonho,RL_bipedal,RL_Dexterous3}. But without a tedious reward-engineering process dedicated to the task at hand, it generally struggles to discover the right behaviors. This can be addressed by leveraging expert demonstrations: multi-modal trajectories can be encoded into motion priors which are then used to guide the training of a robust RL agent \cite{ImitateAndRepurpose,AMP,MultiAMP} in complex loco-manipulation settings. However, instead of relying on expert trajectories generated by human examples \cite{IL_VirtualRealityTeleop,DoorOpeningScienceRobotics}, one could potentially use our framework as an automatic provider of physically-consistent demonstrations for an RL pipeline.

Finally, full autonomy would naturally require incorporating exteroceptive information into our framework. This is necessary to accurately detect and localize the object of interest at test time, as well as to adequately react against unanticipated effects that cannot be inferred with pure proprioception (such as unknown obstacles, or loss of contact with the object). In addition, perception can be used to infer important properties about the object, such as its dimensions, articulation, and affordances, thereby fully replacing user-specified inputs. In the Supplementary Discussion (\textit{Towards real-world deployment}), we discuss the practical usability of our current setup in real-world settings, despite our limiting assumption regarding a-priori-modeled environments. We also propose straightforward extensions that mildly integrate perception with our framework to facilitate deployment.

\section*{Materials and Methods}

\noindent This section provides a detailed description of the architecture depicted in \cref{fig:figure1} with a specific focus on the offline planning module. We justify the importance of different components in the Supplementary Methods (\textit{Ablation study over framework components}).

\subsection*{General optimization formulation}
\noindent The discovery of an optimal plan ${\bm \zeta^*(t)}$ composed of a continuous state trajectory ${\bm x^*(t) \in \mathbb{R}^{n_x}}$ and input trajectory ${\bm u^*(t) \in \mathbb{R}^{n_u}}$ jointly with a sequence of discrete modes ${\bm z^*(t) \in \mathcal{Z} \subset \mathbb{R}^{n_z}}$, can be generically formulated as a mixed-integer optimal control problem \cite{MIOCP}:  
\begin{align} \label{eq:OCProblem}
    \begin{split}
        \bm \zeta^* = \ &\underset{\bm \zeta}{\text{arg\,min}} \ \displaystyle{\int_{0}^{t_f}} L\left(\bm x, \bm u, \bm z\right)dt\\
        &\text{subject to} \quad \dot{\bm{x}} = \bm f_{\bm z}(\bm x, \bm u), \quad \bm x(0) = \bm x_{\text{init}} \\     
        &\qquad \qquad \quad \langle{\bm x, \bm u\rangle} \in \mathcal{S}_{\bm z}, \quad \bm x(t_f) \in \mathcal{X}_{\text{goal}}
    \end{split}
\end{align}
where ${\mathcal{S}_{\bm z} := \Set{\langle{\bm x, \bm u\rangle}}{\bm g_{\bm z}(\bm x, \bm u) = \bm 0, \ \bm h_{\bm z}(\bm x, \bm u) \geq \bm 0}}$ is a set of feasible continuous states and inputs satisfying mode-dependent constraints, and $\bm f_{\bm z}(\bm x, \bm u)$ represents the mode-dependent system dynamics. For the tasks of interest in this paper, the main objective is always specified in terms of a terminal set constraint that is based on a goal region ${\mathcal{X}_{goal}}$ and an unknown final time $t_f$, whereas the cost function ${L(\bm x, \bm u, \bm z)}$ consists of task-independent terms that simply penalize large inputs, unnecessary mode-switching, and deviations from desired nominal configurations. This effectively amounts to solving hybrid constraint satisfaction problems analogous to those arising in the TAMP domain \cite{TAMP_Garrett}. A primary advantage of such a formalism is that it rids us of a cumbersome cost engineering process that would otherwise be required for every newly introduced scenario. 

On the other hand, handling a free end-time mixed-integer nonlinear optimization with a sparse objective using off-the-shelf numerical solvers is computationally intractable. The branching factor involved in standard combinatorial techniques such as branch-and-bound would scale poorly with respect to the number of discrete modes, the time horizon, and the number of time steps used when discretizing the dynamics \cite{METHODS_NumericalOptimalControl}. Alternatively, we propose a formulation that loosens the original problem in \cref{eq:OCProblem} by identifying and exploiting the special structure emerging in multicontact loco-manipulation planning.

Firstly, considering that manipulation modes mostly remain constant over multiple consecutive time steps, we divide the full trajectory into coarse intervals of mode-invariant phases. This allows mode switching to occur less frequently along the horizon, thereby reducing the number of discrete decision variables. Secondly, we introduce an alternative representation by substituting the set of integer variables in \cref{eq:OCProblem} with an equivalent set of contact states ${\bm s \in \mathcal{S}}$ and contact switching actions ${\bm a \in \mathcal{A}}$. Such a representation enables us to define a transition map ${\bm \Gamma: \ \mathcal{S} \times \mathcal{A} \rightarrow \mathcal{S}}$, along with a set of feasible discrete actions $\Bar{\mathcal{A}} \subseteq \mathcal{A}$ constructed using domain-specific logic rules. Applying these modifications yields a more manageable ``mixed-logic program'' whose structure can be leveraged for a fast and effective pruning of infeasible branches. We cast this program in the form of a bilevel optimization consisting of a graph search over discrete state-action sequences interleaved with a contact-driven trajectory optimization that is parameterized by the manipulation modes. The minimization is framed as follows:                      
\begin{equation}\label{eq:BilevelOptimization}
    \begin{split}
        &\underset{\{ \bm s_k, \bm a_k, \bm x_{_{T_k}} \}}{\min} \ \sum_{k=0}^{N-1} E\left(\bm{s}_{k}, \bm{a}_{k}, \bm x_{_{T_k}}\right) \\
        & \quad \ \text{s.t.} \quad  
        \begin{array}[t]{l}
            \langle{\bm{s}_{0}, \ \bm{x}_{_{T_0}}\rangle} = \langle{\bm{s}_{\text {init}}, \ \bm{x}_{\text{init}}\rangle}, \quad \bm{x}_{_{T_N}}\in \mathcal{X}_{\text{goal}} \\[1.5ex] 
            \bm s_{k+1} = \bm \Gamma(\bm s_k, \bm a_k), \quad \bm{a}_{k} \in \Bar{\mathcal{A}}\left(\bm{s}_{k}, \bm{s}_{k-1}, \bm x_{_{T_k}}\right) \\[1.5ex]
            \bm{x}_{_{T_{k+1}}} \in \left\{
                \begin{split}
                    &\underset{\bm{x}(\cdot), \bm{u}(\cdot)}{\text{arg\,min}} \  \Phi\left(\bm x_{_T}, \Tilde{\bm r}\right) + \displaystyle{\int_{0}^{T}}{\mathcal{L}\left(\bm x, \bm u, \Tilde{\bm r}\right)dt} \\[1ex] 
                    &\text{s.t.} \quad  
                    \begin{array}[t]{l} 
                        \bm x(0) = \bm x_{_{T_k}}, \\[0.5ex] 
                        \dot{\bm{x}} = \bm f\left(\bm x, \bm u \mid \bm s_k, \bm a_k\right), \\[0.5ex] 
                        \bm g\left(\bm x, \bm u, t \mid \bm s_k, \bm a_k\right) = \bm 0, \\[0.5ex] 
                        \bm h\left(\bm x, \bm u, t \mid \bm s_k, \bm a_k\right) \geq \bm 0,
                    \end{array} 
                \end{split}
            \right.        
        \end{array}
    \end{split} 
\end{equation}
where $N$ is the unspecified number of directed edges connecting the starting node $\langle{\bm{s}_{0}, \bm{x}_{_{T_0}}\rangle}$ to the goal region $\mathcal{X}_{goal}$. Each edge connecting a node $\langle \bm s_k, \bm x_{_{T_k}} \rangle$ to its neighboring node $\langle \bm s_{k+1}, \bm x_{_{T_{k+1}}} \rangle$ is assigned a transition cost $E$ including both discrete and continuous terms. The continuous component is deduced from the solution of a mode-invariant OCP with a fixed time horizon $T$ and an optimal terminal state $\bm x_{_{T_{k+1}}}$. The inner-level OCP minimizes an objective consisting of an intermediate cost $\mathcal{L}$ and a terminal cost $\Phi$ driven by a reference ${\Tilde{\bm r}\in \mathcal{R} \subseteq \mathbb{R}^{n_x}}$, and is subject to the dynamics and path constraints which are conditioned on a fixed contact mode $\langle \bm s_k, \bm a_k\rangle$. The different elements in the above formulation will be discussed in detail in the remaining subsections. 

\subsection*{Robot-object interaction rules}
\noindent From a user-defined collection of $n_e$ robot end-effectors and $n_c$ potential object contacts, we construct a set of contact states ${\mathcal{S} := \{0, 1, ..., n_c\}^{n_e}}$ that encodes all possible interaction combinations. For instance, a state $\bm s = (0 \ \ 0 \ \ 2)$ indicates that the first two end-effectors are open contacts, and the third one is interacting with the second object contact. Regarding our action space $\mathcal{A}$, we aim to define a minimal set of admissible contact switches since the number of expanded edges in the graph is primarily dictated by the size of that set. To this end, we restrict every action to affect one limb at a time such that it can either keep the same contact state (open/closed), break a closed contact, or establish a new one. Indeed, similar strategies have been previously adopted for acyclic locomotion planning \cite{SamplingBased_EfficientAcyclic} and dexterous manipulation \cite{GraphBased_TrajectoTree}. This rule already enables us to adopt a compact representation for contact switching actions ${\bm a \in \left\{0, 1, ..., n_e\} \times \{0, 1, ..., n_c\right\}}$ where the first and second elements correspond to the limb and object contact indices, respectively. In other words, an action ${\bm a = (e \ \ c)}$ indicates that limb $e$ should establish a new object contact corresponding to index $c$. Moreover, the end-effector $e$ should break its current contact if ${c = 0}$, while the same contact state is maintained if ${e = 0}$. Accordingly, the definition of the transition map $\bm \Gamma$ readily follows as: 
\begin{equation}
    s_{k+1}[i] = 
    \left\{
    \begin{split}
        &s_{k}[i] \qquad   \text{ if } i \neq a_k[1] \qquad \forall i = 1,\ldots,n_e \\
        &a_k[2]   \qquad   \text{if } i = a_k[1] \qquad \forall i = 1,\ldots,n_e \\
    \end{split}    
    \right.
\end{equation}

The branching factor can be further reduced with additional logic rules that make use of loco-manipulation domain knowledge. A central component behind this is the manual classification of robot-object contacts based on certain key attributes: End-effectors are categorized as prehensile/nonprehensile arm contacts, or prehensile/nonprehensile feet that participate in both manipulation and locomotion, whereas object contacts are categorized as prehensile/nonprehensile points or nonprehensile surfaces. Consequently, we can restrict the space of admissible actions $\Bar{\mathcal{A}}$ by introducing a set of basic rules (both generic and domain-specific): For instance, an end-effector is allowed to establish a new object contact only from a previously open contact state; a point object contact can only be occupied by a single limb; and nonprehensile robot contacts cannot be paired with prehensile object contacts. The full list of such pruning rules is provided in the Supplementary Methods (\textit{Loco-manipulation logic rules}).

\subsection*{Contact-driven trajectory optimization}
\noindent When enumerating all possible contact switches, the above conditions only provide us with a fast feasibility checker that marks which edges cannot be expanded. However, an additional step is required to ensure that the admissible modes are also kinematically and dynamically consistent. This can be achieved by adopting sufficiently complex dynamical models in the inner-level optimization of \cref{eq:BilevelOptimization}. We base the OCP formulation in this paper on our previous work \cite{UnifiedMPC}, wherein any arbitrary mobile manipulator is treated as a multi-limbed and poly-articulated floating-base system. Unlike most multi-contact planners that rely on highly simplified kinematic or quasi-static models \cite{GraphBased_LocoManipulationReachability,SamplingBased_EfficientAcyclic,SamplingBased_Edo,SamplingBased_FullDoorOpening,SamplingBased_ContactModeRRTDexterous}, the core idea is to choose a minimal yet high-fidelity model description that captures the dominant coupling effects between the robot's base, its limbs, and the manipulated object. Such a representation can be attained using the robot's full centroidal dynamics \cite{Orin} and a first-order kinematic model augmented with the object's full dynamics. Consequently, we define the continuous state and input as follows: ${\bm x := (\bm x_r \ \ \bm x_o) = (\bm h_{com} \ \ \bm q_b \ \ \bm q_j \ \ \bm q_o \ \ \bm v_o) \text{ and } \bm u = (\mathbf{w}_e \ \ \bm v_j)}$. The robot state $\bm x_r$ consists of the centroidal momentum $\bm h_{com}$, base pose $\bm q_b$, and joint positions $\bm q_j$, whereas the object state $\bm x_o$ comprises its generalized coordinates $\bm q_o$ and velocities $\bm v_o$. The input $\bm u$ stacks the contact wrenches $\mathbf{w}_e$ acting at the end-effectors and the joint velocities $\bm v_j$. It is important to note that the inclusion of the manipulandum's dynamics in the Equations of Motion (EoM) presumes a reasonable knowledge of the object's nominal model, in addition to its kinematic and dynamic parameters. Moreover, in contrast to our previous formulation in \cite{UnifiedMPC}, the set of end-effector contact forces appearing in the object dynamics is now conditioned on the manipulation mode. A complete description of the EoM is provided in the Supplementary Methods (\textit{Contact-driven OCP formulation: Dynamics}).        

The central design principle underlying our framework is one that aims to capture various loco-manipulation behaviors in a unified manner by predominantly relying on a constraint-based formulation while keeping the cost engineering process as lightweight as possible. Due to the generic contact classification discussed earlier, the same constraint set would be applicable to diverse scenarios involving arbitrary mobile manipulators or articulated objects. Besides standard continuously-active constraints enforcing system operational limits, workspace limits, and collision-avoidance \cite{AlmaSelfCollision}, we introduce a collection of contact-driven constraints that are parameterized by the manipulation mode. The mode can be readily extracted from the current contact state $\bm s_k$ and the contact switching action $\bm a_k$. For instance, a closed and prehensile contact state would entail a zero relative twist condition between the gripper and the graspable object contact. In contrast, a nonprehensile interaction (wherein only sticking contacts are assumed) would only constrain their relative linear velocity while ensuring the end-effector force lies within the friction cone. Furthermore, we preserve the same time-based switched system structure adopted in \cite{TO_Farbod,UnifiedMPC} when defining our constraints, hence the time dependence in the path constraints of \cref{eq:BilevelOptimization}. By this, we are able to encode certain hybrid behaviors that can be described with basic mode schedules: a fixed mode sequence coupled with a set of switching times. These could include cyclic gaits which are primarily relevant for legged mobile manipulators, or simple manipulation schedules such as breaking a closed contact, or establishing a new contact. An elaborate description of the OCP constraints is presented in the Supplementary Methods (\textit{Contact-driven OCP formulation: Constraints}).     

The definition of the OCP cost function consisting of $\mathcal{L}$ and $\Phi$ is straightforward, as it only encompasses simple quadratic terms that mostly use the same set of weighting matrices for different tasks. As described in the Supplementary Methods (\textit{Contact-driven OCP formulation: Cost Function}), they include elements that penalize large inputs and regularize the robot's state around a nominal configuration, as well as a task-dependent term that takes care of tracking target references $\Tilde{\bm r}$ corresponding to both the object and the robot's base.  

\subsection*{Offline planner: Sampling-based bilevel optimization}
\noindent In order to address the bilevel optimization in \cref{eq:BilevelOptimization}, we start by noting that we do not encode our loco-manipulation tasks in the form of a terminal symbolic state $\bm s_N$, but rather as a goal region to which a reachable continuous state $\bm x_{_{T_N}}$ must belong. This would typically require augmenting the original graph state (namely, the contact state) with a discretization of the continuous space \cite{TAMP_DyadicManipulation}. Alternatively, instead of inefficiently searching through a dense graph of high-dimensional augmented states, we resort to an approximate sampling-based solution to the bilevel search. Such a scheme enables a rapid computation of feasible solutions and offers an effective globalization property that helps avoid bad local minima. Although this happens at the expense of optimality, our primary target is nonetheless fulfilled: discovering a physically consistent sub-optimal plan that solves the task.       

To elaborate, the planning algorithm operates by incrementally building a multi-modal tree in which each tree node is characterized by its contact mode and a reachable terminal state extracted from the optimal trajectory of the corresponding edge. The tree expansion is guided by two key components: a node selection mechanism and a strategy that determines the cost targets $\Tilde{\bm r}$ during node extensions. Similar in essence to \cite{FrazzoliAutonomousDriving}, the latter relies on a sampling-based approach that drives our inner-level OCP with targets defined in a smaller subset of the full state space, namely the reference space ${\mathcal{R}}$. Specifically, in our case, this corresponds to the robot's 2D base pose and the object's generalized coordinates. In fact, the goal region itself is constructed in terms of our reference space (e.g., proximity to a target base position or object configuration), rendering the task-specification step notably simpler for the user. As described in the Supplementary Methods (\textit{Reference generation}), the cost reference ${\Tilde{\bm r} := \Tilde{\bm r}(\bm s,\bm a, \bm x_{_T}, \Tilde{\bm y}) \in \mathcal{R}}$ basically determines the direction of extension and is generated as a function of the contact mode, the node's continuous state, and a random variable ${\Tilde{\bm y} \in SE(2)\times\mathbb{R}^{n_o}}$. Whilst considering the problem's workspace limits, we sample $\Tilde{\bm y}$ either from a uniform distribution $\mathcal{U}$ or a goal-directed normal distribution $\mathcal{N}$ whose mean is used in constructing the goal set $\mathcal{X}_{goal}$.

Regarding the outer-level selection step, each tree node is first assigned a cumulative cost ${C_n = \sum_{i=0}^{n-1} E_i + \alpha H}$ that is composed of the accumulated edge-costs from the starting node and a heuristic function $H$. The one with the lowest cost $C_n$ is then chosen for expansion using an Anytime Non-parametric ${\text{A}^\text{*}}$ (${\text{ANA}^\text{*}}$) strategy \cite{ANA_Star}. This method aims for a quick discovery of feasible plans by initially applying a best-first selection criterion, then keeps enhancing the solution as it converges towards an ${\text{A}^\text{*}\text{-type}}$ search by gradually decreasing the weight $\alpha$ to one. Each edge cost combines a discrete element penalizing contact switches with a merit function composed of continuous regularization terms and penalties on constraint violations in the corresponding OCP. Moreover, we evaluate our heuristic cost as ${H=N_{segments}\cdot E_{average}}$ where $N_{segments}$ roughly estimates the number of unconstrained trajectory segments required by the robot/object to reach the goal assuming it can move freely at maximum velocity, whereas $E_{average}$ denotes the average edge cost of all branches in the existing tree. The aim with such a choice is to ultimately converge toward a near-optimal solution by approaching an admissible heuristic function that underestimates the true cost-to-go.  

The pseudo-code for an elementary version of the algorithm used to solve the bilevel optimization in \cref{eq:BilevelOptimization} is presented below:\\[2ex]
\noindent\makebox[\linewidth][c]{%
\begin{minipage}{0.9\linewidth}
\textbf{Algorithm 1:} Multi-Contact Loco-Manipulation Planner\\
\null \ Initialize $\bm s_{init}$, $\bm x_{init}$, $\mathcal{X}_{goal}$, $\mathcal{N}$, $\mathcal{U}$, $T$, $\alpha$ \\
\null \ Define tree $\mathcal{T}$, set of open nodes $\mathcal{O}$, and solution set $\mathcal{S}^*$ \\
\null \ Initialize start node $n_s$ using initial states and append to $\mathcal{T}$\\
\null \ \textbf{while} not termination condition reached \textbf{do}\\
    \null \quad \texttt{/* Goal directed extension */}\\
    \null \quad Draw sample $\Tilde{\bm y}$ from $\mathcal{N}$\\
    \null \quad \textbf{while} not $\mathcal{O}$ is empty and not max iterations reached \textbf{do}\\
        \null \qquad Set node last appended to $\mathcal{T}$ as initial node $n_{i}$\\
        \null \qquad \textbf{if} $n_i$ state belongs to $\mathcal{X}_{goal}$ \textbf{then}\\
            \null \qquad \quad Append $n_i$ to $\mathcal{S}^*$; update weight $\alpha$; break\\
        \null \qquad \textbf{end if} \\
        \null \qquad Extend $n_i$ and update $\mathcal{O}$, $\mathcal{T}$: GenerateSuccessors($n_i$, $\Tilde{\bm y}$) \\             
    \null \quad \textbf{end while}\\
    \null \quad \texttt{/* Uniformly random extension */} \\
    \null \quad \textbf{while} not new node added to tree \textbf{do}\\
        \null \qquad Draw sample $\Tilde{\bm y}$ from $\mathcal{U}$\\
        \null \qquad Set node in $\mathcal{T}$ nearest to $\Tilde{\bm y}$  as $n_i$ using a proper metric\\
        \null \qquad Extend $n_i$ and update $\mathcal{O}$, $\mathcal{T}$: GenerateSuccessors($n_i$, $\Tilde{\bm y}$) \\             
    \null \quad \textbf{end while}\\        
\null \ \textbf{end while} 
\end{minipage}}\\[2ex]

\noindent\makebox[\linewidth][c]{%
\begin{minipage}{0.9\linewidth}
\textbf{Algorithm 2:} Function to Generate Successors in Algorithm 1\\
\null \ \textbf{function} GenerateSuccessors($n_i$, $\Tilde{\bm y}$) \\
    \null \quad \textbf{for each} $\bm a \in \Bar{\mathcal{A}}(n_i)$ \textbf{do}\\            
        \null \qquad Compute OCP references $\Tilde{\bm r}(n_i, \bm a, \Tilde{\bm y})$\\
        \null \qquad Solve OCP given $n_i$, $\bm a$, $\Tilde{\bm r}$, and the time horizon $T$\\
        \null \qquad \textbf{if} OCP solution found \textbf{then}\\
            \null \qquad \quad Set state and cost info of new node $n_f$ from solution\\
            \null \qquad \quad \textbf{if} cost from $n_s$ to $n_f <$ smallest cost in $\mathcal{S}^*$ \textbf{then}\\
                \null \qquad \qquad Append $n_f$ to $\mathcal{O}$\\                
            \null \qquad \quad \textbf{end if} \\
        \null \qquad \textbf{end if} \\
    \null \quad \textbf{end for}\\
    \null \quad Select node in $\mathcal{O}$ with least cumulative cost and append to $\mathcal{T}$\\  
\null \ \textbf{end function}
\end{minipage}}\\[2ex]
\noindent In summary, the above algorithm boils down to an alternation between goal-directed tree expansions and an RRT-like exploration of the reference space. Indeed, the core component driving our planner is a targeted balance between exploration and exploitation. This aspect was fundamental in choosing an appropriate nearest-neighbor metric and in our introduction of supplementary pruning strategies that help maintain a sparse yet sufficiently rich tree. Implementation details regarding the underlying OCP solver, the exploration metric, and the pruning process can be found in the Supplementary Methods (\textit{Implementation details}), whereas the common hyperparameters used for all tasks are given in \cref{tables4}.

\subsection*{Offline planner: Long-horizon trajectory optimization}
\noindent A complete plan connecting a sequence of smooth trajectories may not necessarily be smooth itself, especially when randomness is involved in generating the plan. Therefore, we introduce a post-processing step that enhances the quality of the solutions obtained from the bilevel search. To this end, we use the trajectory sequence as a feasible initial guess to warm-start a long-horizon trajectory optimization routine. Although not required, the sequence can also be applied as a cost reference $\Tilde{\bm r}$ to guide the optimization. The resulting OCP structure resembles that in \cref{eq:BilevelOptimization}, but at this stage, the short time horizon $T$ is replaced with the total duration ${N \cdot T}$ needed to solve the loco-manipulation task. Furthermore, the state-action pairs governing the path constraints over the $N$ time segments are now predetermined from the discovered contact schedule. 

Apart from smoothening the full trajectory and getting rid of superfluous motions, the long-horizon optimization offers extra benefits with regard to how mode switches are handled. Unlike the contact-dependent OCPs in \cref{eq:BilevelOptimization}, the added lookahead capabilities result in refined solutions that suitably alter contact locations during mode switching while accounting for future events and maneuvers (e.g., adapted foothold locations or adapted end-effector positions when interacting with a surface contact). Ultimately, this leads to smaller contact forces and more robust behaviors associated with increased stability and higher chances for successful task attainment.      
\subsection*{Two-layer tracking controller}
\noindent When it comes to executing the offline multi-contact plans on the real robot, a pointwise tracking scheme would normally be sufficient for quasi-static motions and statically stable systems. Since this does not hold in our case of a quadrupedal mobile manipulator performing dynamic loco-manipulation tasks, we resort to a two-layer tracking architecture providing stronger robustness properties. A schematic diagram detailing the interconnections between the two layers is provided in \cref{figs4}. 

The first layer is a whole-body model predictive controller \cite{UnifiedMPC} that acts as an effective filter with disturbance-rejection properties. Incorporating this component is instrumental in achieving behaviors involving multiple contact switches. Otherwise, such phenomena would induce unmodelled effects that could render the reference trajectory unstabilizable with a simple one-step lookahead strategy. In contrast to \cite{UnifiedMPC}, the MPC in this work is perceived as a tracker of precomputed physically-consistent plans rather than a planner that generates new solutions for object-centric or robot-centric objectives. With such a perspective, we manage to keep the MPC formulation as elementary as possible by removing the object state $\bm x_o$, reducing the number of constraints, and adopting simple state-input quadratic cost functions driven by the robot's offline references ${\langle\bm x^*_r, \bm u^*\rangle}$. As a result, we obtain a task-independent MPC-based tracker with a considerable gain in its computational rate. The adopted MPC cost weights are shown in \cref{tables6}.        

The second layer is a whole-body reactive controller that outputs joint position-velocity-torque commands to track the higher-level MPC references. The references are first mapped into desired generalized coordinates and velocities, using conversions between the full centroidal dynamic model and the full rigid-body dynamics \cite{UnifiedMPC}. A series of cascaded quadratic programs are then solved in a hierarchical fashion based on a list of prioritized tasks \cite{DarioWBC1,DarioWBC2}. This results in optimal generalized accelerations and contact forces that are used to compute reference joint torques from the inverse dynamics. A list of the tasks included in the hierarchical QP along with their corresponding priorities is provided in \cref{tables7}.    

\section{Acknowledgments}
We would like to thank Jan Preisig, Simone Arreghini, and Jia-Ruei Chiu for their valuable help in developing the autonomous door traversal pipeline and their help with hardware experiments. \textbf{Funding:} The project was funded, in part, by the Intel Network on Intelligent Systems, the Swiss National Science Foundation (SNF) through the National Centre of Competence in Research Robotics (NCCR Robotics), the European Research Council (ERC) under the European Union's Horizon 2020 research and innovation programme grant agreement No 852044, the Swiss National Science Foundation through the National Centre of Competence in Digital Fabrication (NCCR dfab), and TenneT TSO. This work has been conducted as part of ANYmal Research, a community to advance legged robotics. \textbf{Author contributions:} J-P.S. formulated and implemented the planning and control architecture, designed and performed all simulations and real-world experiments, and wrote the manuscript. F.F. contributed to the development of the planning algorithm and the design of the experiments. F.F. and M.H. contributed to the analysis of the data, revised the manuscript, and refined ideas. \textbf{Competing interests:} The authors declare that they have no competing interests. \textbf{Data and materials availability:} All data needed to evaluate the conclusions in the paper are present in the paper or the Supplementary Materials. Other materials can be found at \href{https://zenodo.org/badge/latestdoi/384033964}{SupplementaryCode} and \href{https://doi.org/10.5061/dryad.vq83bk3zp}{SupplementaryData} (DOI: \textit{doi:10.5061/dryad.vq83bk3zp})

\section*{Supplementary materials}
\makebox[1.8cm][l]{Section S1.} Nomenclature \\
\makebox[1.8cm][l]{Section S2.} Contact-driven OCP formulation: Dynamics \\
\makebox[1.8cm][l]{Section S3.} Contact-driven OCP formulation: Constraints \\
\makebox[1.8cm][l]{Section S4.} Contact-driven OCP formulation: Cost Function \\
\makebox[1.8cm][l]{Section S5.} Reference generation \\
\makebox[1.8cm][l]{Section S6.} Implementation details \\
\makebox[1.8cm][l]{Section S7.} Ablation study over framework components \\
\makebox[1.8cm][l]{Section S8.} Towards real-world deployment \\
\makebox[1.8cm][l]{Figure S1.} Robot-Object reference frames.\\
\makebox[1.8cm][l]{Figure S2.} Robot-Object collision bodies.\\
\makebox[1.8cm][l]{Figure S3.} Ablation study over framework components.\\
\makebox[1.8cm][l]{Figure S4.} Block diagram detailing MPC-WBC interconnections.\\
\makebox[1.8cm][l]{Table S1.} Objects model parameters. \\
\makebox[1.8cm][l]{Table S2.} Inner-Level OCP cost weights. \\
\makebox[1.8cm][l]{Table S3.} Task-dependent hyperparameters for \\ 
\makebox[1.8cm][l]{}  reference generation. \\
\makebox[1.8cm][l]{Table S4.} Task-independent hyperparameters for bilevel search. \\
\makebox[1.8cm][l]{Table S5. } Task-independent hyperparameters for NNS metric \\
\makebox[1.8cm][l]{} and auxiliary strategies. \\
\makebox[1.8cm][l]{Table S6. } MPC cost weights. \\
\makebox[1.8cm][l]{Table S7. } WBC prioritized tasks list. \\
\makebox[1.8cm][l]{Movie S1.} Generated behaviors for dishwasher manipulation task.\\ 
\makebox[1.8cm][l]{Movie S2.} Generated behaviors for valve manipulation task. \\ 
\makebox[1.8cm][l]{Movie S3.} Generated behaviors for movable obstacle traversal task. \\
\makebox[1.8cm][l]{Movie S4.} Generated behaviors for door traversal task.\\
\makebox[1.8cm][l]{Movie S5.} Ablation study over framework components. \\
\makebox[1.8cm][l]{Movie S6.} Demonstration of inherent robustness.\\
\makebox[1.8cm][l]{Movie S7.} Towards autonomous door traversal. \\

\bibliography{scibib}

\begin{thebibliography}{100}
\bibitem{SpotWithArm}
New dog-like robot from boston dynamics can open doors.
\end{thebibliography}

\clearpage
\newpage

\setcounter{table}{0}
\makeatletter 
\renewcommand{\thetable}{S\@arabic\c@table}
\makeatother

\setcounter{figure}{0}
\makeatletter 
\renewcommand{\thefigure}{S\@arabic\c@figure}
\makeatother

\setcounter{algorithm}{0}
\makeatletter 
\renewcommand{\thealgorithm}{S\@arabic\c@algorithm}
\makeatother

\subsection*{S1. Nomenclature}
\makebox[2.25cm]{$\mathcal{I}, \mathcal{O}, \mathcal{B}, \mathcal{G}   $} Inertial, object, base, and centroidal reference  \\
\makebox[2.25cm]{}
frames, respectively \\
\makebox[2.25cm]{$n_j, n_o$} Number of actuated robot joints, and object degrees\\
\makebox[2.25cm]{} of freedom, respectively \\
\makebox[2.25cm]{$n_e, n_c $} Number of robot end-effectors, and object contacts, \\
\makebox[2.25cm]{} respectively \\
\makebox[2.25cm]{$e_i, c_i           $} End-effector contact $i$, and object contact $i$,\\
\makebox[2.25cm]{} respectively \\
\makebox[2.25cm]{$\mathcal{E}_i, \mathcal{C}_i$} Reference frames attached to $e_i$, and $c_i$, respectively \\
\makebox[2.25cm]{$\bm x, \bm u$} Continuous robot-object state, and input, \\
\makebox[2.25cm]{} respectively\\
\makebox[2.25cm]{$\bm s, \bm a$} Discrete contact state, and contact-switching action,\\
\makebox[2.25cm]{} respectively\\
\makebox[2.25cm]{$\bm q_b$} Floating-base pose (position and orientation  as \\
\makebox[2.25cm]{} ZYX- Euler Angles)\\
\makebox[2.25cm]{$\bm q_j, \bm v_j$} Joint positions, and velocities, respectively\\
\makebox[2.25cm]{$\bm q_r, \bm q_o$} Robot, and object generalized coordinates, \\
\makebox[2.25cm]{} respectively\\
\makebox[2.25cm]{$\bm v_r, \bm v_o$} Robot, and object generalized velocities,\\
\makebox[2.25cm]{} respectively\\
\makebox[2.25cm]{$\bm x_r, \bm x_o$} Robot, and object continuous state, respectively\\
\makebox[2.25cm]{$\bm C_{A,B}$} Rotation matrix for orientation of frame $\{\mathcal{B}\}$ w.r.t.  \\
\makebox[2.25cm]{} frame $\{\mathcal{A}\}$ \\
\makebox[2.25cm]{$\bm f_{e_i}, \bm \tau_{e_i}, \mathbf{w}_{e_i}$} Force, torque, and wrench at contact $e_i$ in $\{\mathcal{I}\}$ \\
\makebox[2.25cm]{} respectively\\
\makebox[2.25cm]{${\bm r}_{B},{\bm \nu}_{B}, {\bm \omega}_{B}  $} Absolute position, linear velocity, and angular \\
\makebox[2.25cm]{} velocity of $B$ in $\{\mathcal{I}\}$, respectively \\
\makebox[2.25cm]{${\bm r}_{A,B},{\bm \nu}_{A,B},{\bm \omega}_{A,B}  $} Relative position, linear velocity, and angular \\
\makebox[2.25cm]{} velocity of $B$ w.r.t. $A$ in $\{\mathcal{S}^c\}$, respectively\\
\makebox[2.25cm]{$\mathcal{S}^e_{open}$} Set of end-effectors with an open contact\\
\makebox[2.25cm]{$\mathcal{S}^e_{ground}$} Set of end-effectors with a closed ground contact \\
\makebox[2.25cm]{} (for locomotion)\\
\makebox[2.25cm]{$\mathcal{S}^e_{object}$} Set of end-effectors with a closed object contact\\
\makebox[2.25cm]{$\mathcal{S}^e_{break}$} Set of end-effectors breaking an object contact\\
\makebox[2.25cm]{$\mathcal{S}^e_{establish}$} Set of end-effectors establishing an object contact\\
\makebox[2.25cm]{$\mathcal{S}^e_{switch}$} Set of end-effectors switching their manipulation \\
\makebox[2.25cm]{} contact state $\left(\mathcal{S}^e_{break} \cup \mathcal{S}^e_{establish}\right)$\\
\makebox[2.25cm]{$\mathcal{S}^e_{active}$} Set of end-effectors switching or with a closed \\
\makebox[2.25cm]{} object contact $\left(\mathcal{S}^e_{switch} \cup \mathcal{S}^e_{object}\right)$ \\
\makebox[2.25cm]{$\mathcal{S}^e_{foot}, \mathcal{S}^e_{arm}$}  Set of end-effectors classified as feet, and  \\
\makebox[2.25cm]{} arm contacts, respectively\\
\makebox[2.25cm]{$\mathcal{S}^e_{preh}, \mathcal{S}^e_{nonpreh}$} Set of end-effectors classified as prehensile, and \\
\makebox[2.25cm]{} nonprehensile, respectively \\
\makebox[2.25cm]{$\mathcal{S}^c_{preh}, \mathcal{S}^c_{nonpreh}$} Set of object contacts classified as prehensile, and \\
\makebox[2.25cm]{} nonprehensile, respectively \\
\makebox[2.25cm]{$\mathcal{S}^c_{point}, \mathcal{S}^c_{surface}$} Set of object contacts classified as points, and \\
\makebox[2.25cm]{} surfaces, respectively

\subsection*{Section S2. Contact-driven OCP formulation: Dynamics}

\begin{figure}[htb!]
\vspace{2cm}
\centering
  \makebox[\linewidth]{
\includegraphics[width=\linewidth, keepaspectratio]{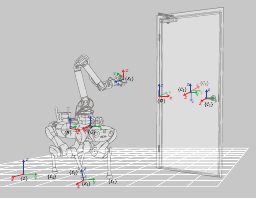}}
  \caption{\textbf{Robot-Object reference frames.} The centroidal frame $\{\mathcal{G}\}$ is attached to the robot’s center of mass and is always aligned with the world (inertial) frame $\{\mathcal{I}\}$. All end-effector frames $\{\mathcal{E}_i\}$ and object contact frames $\{\mathcal{C}_i\}$ are defined using a specific convention to ensure that the same OCP formulation can be applied to any loco-manipulation scenario.}
  \label{figs1} 
\end{figure}

We start by noting that whenever a vector quantity is used without any reference frame associated to it, then it is automatically assumed to be expressed in the inertial frame $\{\mathcal{I}\}$. This holds true for all upcoming sections as well. Moreover, we refer the reader to \cref{figs1} for an illustration of the modeled system and the different reference frames used in our formulation.       

We devise the system flow map $\dot{\bm x} = \bm f\left(\bm x, \bm u \mid \bm s, \bm a\right)$ in the OCP using the Equations of Motion (EoM) derived from the robot's centroidal dynamics, its first-order kinematics, and the full rigid-body dynamics of an arbitrary object being manipulated by the robot:

\begin{equation}
    \begin{cases}
        \dot{\bm p}_{com} = \displaystyle \sum\limits_{i = 1}^{n_e} \bm f_{e_i} + m \bm g \\
        \dot{\bm l}_{com} = \displaystyle \sum\limits_{i = 1}^{n_e} \bm r_{com, e_i}(\bm q_r) \times \bm f_{e_i} + \bm \tau_{e_i}\\
        \dot{\bm q}_b = \bm A^{-1}_b(\bm q_r) \left(\bm h_{com} - \bm A_j(\bm q_r) \dot{\bm q}_j \right) \\   
        \dot{\bm q}_j = \bm v_j \\
        \dot{\bm q}_o = \bm v_o \\
        \dot{\bm v}_o = \bm M^{-1}_o(\bm q_o) \left(\underset{e_i \in \mathcal{S}^e_{active}}{\displaystyle {\sum}}{_\mathcal{O}\bm J^T_{o,e_i}}(\bm q_o,\bm q_r) \ _\mathcal{O}{\mathbf{w}_{e_i}} - \bm b_o(\bm q_o, \bm v_o) \right)
    \end{cases}
\end{equation}

where the centroidal momentum $\bm h_{com} = (\bm p_{com} \ \ \bm l_{com}) \in \mathbb{R}^{6}$ is composed of the linear momentum $\bm p_{com}$ and the angular momentum $\bm l_{com}$ about the centroidal frame $\{\mathcal{G}\}$. A key element in the EoM is the Centroidal Momentum Matrix $\bm A(\bm q_r) = \begin{bmatrix} \bm A_b \ \ \ \bm A_j \end{bmatrix} \in \mathbb{R}^{6\times(6+n_j)}$ which maps the time-derivative of the generalized coordinates $\dot{\bm q}_r$ to the centroidal momentum. It enables us to capture the influence of the limbs' motion on the floating-base dynamics, which is particularly essential when the mass of the limbs constitutes a considerable portion of the total whole-body mass $m$. In fact, alternative template models adopted in the legged locomotion literature such as the Single Rigid Body Dynamics (SRBD) model \cite{MitCheetah,FeedbackMPCRuben,TO_Winkler} are basically simplified variants of the Full Centroidal Dynamics (FCD) model presented above. Therefore, they can be similarly applied in the context of our framework depending on the task requirements (e.g., fast dynamic motions vs. slow quasi-static motions) and the mobile manipulator platform being used (e.g., heavy limbs vs. massless limbs assumption).

As for the components defining the manipulandum dynamics, we denote the object's generalized mass matrix by $\bm M_o(\bm q_o)$ and the generalized state-dependent effects (e.g., recoil, damping, static friction, etc.) by $\bm b_o(\bm q_o, \bm v_o)$. Unlike our previous formulation in \cite{UnifiedMPC} where a single and prespecified interaction location was assumed, the generalized external torque is now determined by a collection of manipulation contacts conditioned on the contact mode. More specifically, the elements of the set $\mathcal{S}^e_{active}$, and accordingly the contact Jacobians ${_{\mathcal{O}}\bm J_{o,e_i}(\bm q_o, \bm q_r) \in \mathbb{R}^{6\times n_o}}$ are all functions of the contact state and action pair $\langle\bm s, \bm a\rangle$. To construct $\mathcal{S}^e_{active}$, we collect the end-effectors in a closed contact state from $\bm s$ and the contact-switching end-effector (if applicable) from $\bm a$. Regarding the computation of $\bm_\mathcal{O}J_{o,e_i}$, we need to identify the right object links and the proper points on these bodies at which we evaluate the contact Jacobians. To this end, we extract the links corresponding to the object contacts $c_i$ that are coupled with the end-effectors $e_i$ from $\bm s$ and $\bm a$ as well, and the contact locations on these links from $_{\mathcal{O}}{\bm r}_{o,e_i}(\bm q_r)$. The model parameters for all objects introduced in this work are given in \cref{tables1}. 
\subsection*{Section S3. Contact-driven OCP formulation: Constraints}
Below, we present the different constraints introduced in the OCP and divide them into two main classes: constraints that are continuously-enforced, and contact-driven constraints that are activated/deactivated based on the manipulation mode deduced from $\bm s$ and $\bm a$.  
\\\\ \textbf{Continuously-enforced constraints}\\
These include standard constraints such as workspace limits for the robot's base and the object, desired bounds on their velocities, in addition to system operational limits imposed on the joints' positions and velocities:
\begin{equation}
\left\{
\begin{split}
&\bm q_r^{min} \leq \bm q_r \leq \bm q_r^{max}\\
&\bm q_o^{min} \leq \bm q_o \leq \bm q_o^{max}\\
&\bm v_r^{min} \leq \bm v_r \leq \bm v_r^{max}\\
&\bm v_o^{min} \leq \bm v_o \leq \bm v_o^{max}\\
\end{split}
\right.
\end{equation}
Another essential constraint takes care of avoiding robot self-collisions as well as collisions between the robot, object, and environment. This is done by constraining the signed distances $\bm d(\bm x_r, \bm x_o) \in \mathbb{R}^{n_p}$ between $n_p$ pairs of links that are represented with primitive collision bodies:  
\begin{equation}
    \bm d(\bm x_r, \bm x_o) - \epsilon \cdot \bm 1_{n_p\times1} \geq \bm 0_{n_p\times1}
\end{equation}
where $\epsilon \geq 0$ is the minimum allowable distance between collision pairs. Similar to our previous work \cite{AlmaSelfCollision}, we query the distances using the Gilbert-Johnson-Keerthi (GJK) algorithm \cite{GJK} provided in the \emph{Flexible Collision Library} (\emph{FCL}) \cite{FCL}.

\begin{figure}[htb!]

    \centering
      \makebox[\linewidth]{
    \includegraphics[width = \linewidth, keepaspectratio]{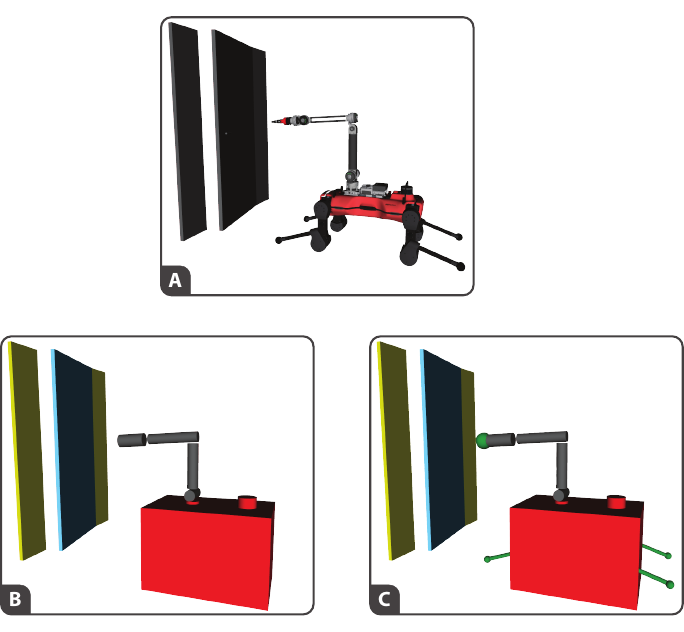}}
    \caption{\textbf{Robot-Object collision bodies.} (\textbf{A}) Visualization of the robot and door at a specific configuration. (\textbf{B}) Collision bodies for continuously-enforced constraints. (\textbf{C}) The collision bodies representing the last link of each limb (colored in green) are only added  into the collision-avoidance constraints when the corresponding limb is establishing or breaking contact with the object.} 
    \label{figs2}
\end{figure}

A major and inevitable issue arises when incorporating (3) within a manipulation planner due to two conflicting objectives: seeking contact between bodies for manipulation (i.e., an end-effector with an object link) while simultaneously ensuring they are collision-free. To resolve this, we split the set of collision bodies into continuously-active components that are included in (3) and contact-dependent components that are only active under certain conditions on the manipulation mode. The latter will be discussed in detail in the \emph{Contact-driven Constraints} subsection. For an illustration of the primitive shapes used to represent the different collision bodies, we refer the reader to \cref{figs2}.   
\\\\ \textbf{Contact-driven constraints}\\
These include constraints that are defined at the level of the end-effector contacts. The activation of each individual constraint is determined by the sets $\mathcal{S}^e$ and $\mathcal{S}^c$ which are constructed based on the contact state $\bm s$, the contact-switching action $\bm a$, and the time $t\in[0,T]$. We recapitulate that the dependence on time is due to the use of predefined mode schedules that capture simple yet generic behaviors over short-horizons such as cyclic gaits, or manipulation modes involving making/breaking an object contact. It is worth noting that the selection of a desired gait is also determined by $\bm s$ and $\bm a$. For instance, one could impose a trotting gait whenever $\bm a$ is a non-switching action, and a stance phase otherwise. All contact-dependent constraints are presented below and are grouped according to their activation conditions.    
\begin{enumerate}
    \item \textbf{Closed object contact} $\left(e_i \in \mathcal{S}^e_{object}\right)$:
        \begin{equation}
        \scalemath{0.8}{
        \left\{
            \begin{split}
            &\bm \nu_{e_i}(\bm q_r) - \bm \nu_{e_i}\left(\bm q_o, \bm r_{e_i}(\bm q_r)\right) = \bm 0_{3\times1} &&\textit{(zero relative linear velocity)} \\
            &_{\mathcal{C}_i}\bm M_l^T \ \bm _{\mathcal{C}_i}\bm f_{e_i} = \bm 0_{m_{l}\times1}  &&\textit{(no forces along motion-constrained directions)}
            \end{split}
        \right.}
        \end{equation}
        where $_{\mathcal{C}_i}\bm M_l \in \mathbb{R}^{3\times m_l}$ is a matrix collecting all directions of the object contact frame ${\mathcal{C}_i}$ (expressed in $\{\mathcal{C}_i\}$) with a constrained linear motion. The second constraint is optional; it was introduced to get rid of any unnecessary internal forces. Moreover, a closed object contact could involve either a nonprehensile or a prehensile interaction, each of which adds its own set of constraints:  
    \begin{enumerate}[label=(\alph*)]
    \item \textbf{Closed \& nonprehensile object contact} $\left(e_i \in \mathcal{S}^e_{object} \land c_i \in \mathcal{S}^c_{nonpreh} \right)$:
        \begin{equation}
        \scalemath{0.8}{
        \left\{
            \begin{split}
            &\bm \tau_{e_i} = 0  && \textit{(zero contact torques)}\\
            &_{\mathcal{C}_i}\hat{\bm n}^T \left(\bm C_{\mathcal{C}_i,\mathcal{E}_i} \cdot \ _{\mathcal{E}_i}\hat{\bm n}\right) \geq 0 && \textit{(end-effector normal inside halfspace)} \\        
            &_{\mathcal{C}_i}\hat{\bm n}^T \ _{\mathcal{C}_i}\bm f_{e_i} \geq 0  && \textit{(unilateral force in pushing direction)} \\
            &\mu_s \cdot \left(_{\mathcal{C}_i}\hat{\bm n}^T \ _{\mathcal{C}_i}\bm f_{e_i}\right) - \left\lVert _{\mathcal{C}_i}\hat{\bm T}^T \ _{\mathcal{C}_i}\bm f_{e_i}\right\rVert_{_2} \geq 0 && \textit{(end-effector force inside friction cone)}            
            \end{split}
        \right.}
        \end{equation} 
        where $_{\mathcal{C}_i}\hat{\bm n}$ and $_{\mathcal{E}_i}\hat{\bm n} \in \mathbb{R}^3$ are the unit normal vectors at contacts $c_i$ and $e_i$, respectively. $_{\mathcal{C}_i}\hat{\bm T} \in \mathbb{R}^{3\times2}$ collects the unit vectors from frame $\{{\mathcal{C}_i}\}$ that are tangent to the contact surface, and $\mu_s$ is the friction coefficient. 
        We note that the third and fourth constraints are not included simultaneously: in the case where the optional constraint of (4) is applied (i.e., no forces along motion-constrained directions), we impose the unilateral constraint, otherwise we enforce the friction cone instead. 
    \item \textbf{Closed \& prehensile object contact} $\left(e_i \in \mathcal{S}^e_{object} \land c_i \in \mathcal{S}^c_{preh} \right)$:         
        \begin{equation}
        \scalemath{0.8}{
        \left\{
            \begin{split}
            &\bm \omega_{\mathcal{E}_i}(\bm q_r) - \bm \omega_{\mathcal{C}_i}\left(\bm q_o\right) = \bm 0_{3\times1} && \textit{(zero relative angular velocity)} \\
            &_{\mathcal{C}_i}\bm M_r^T \ \bm _{\mathcal{C}_i}\bm \tau_{e_i} = \bm 0_{m_{r}\times1} && \textit{(no torques around motion-constrained directions)}
            \end{split}
        \right.}
        \end{equation}
        where $_{\mathcal{C}_i}\bm M_r \in \mathbb{R}^{3\times m_r}$ is a matrix collecting all directions of the object contact frame ${\mathcal{C}_i}$ (expressed in $\{\mathcal{C}_i\}$) with a constrained rotational motion.
    \end{enumerate}
    \item \textbf{Open contact} $\left(e_i \in \mathcal{S}^e_{open} \right)$:    
        \begin{equation}
        \left  
        \{
        \begin{split}
        &\mathbf{w}_{e_i} = 0  && \textit{(zero contact wrench)} \\
        &\bm d(\bm x_r, \bm x_o) - \epsilon \cdot \bm 1_{m_p\times1} \geq \bm 0_{m_p\times1}  &&  \textit{(collision avoidance)}
        \end{split}
        \right.
        \end{equation} 
        where $\bm d \in \mathbb{R}^{m_p}$ is a vector of signed distances between new collision pairs (i.e., additional to those in the \emph{Continuously-enforced constraints}) generated by accounting for collision bodies at the robot's end-effectors, as shown in \cref{figs2}. This is particularly important during contact switching as the robot is in close proximity to the object right before or after a closed contact state. Moreover, a contact-breaking and contact-making action both involve an open-contact phase, each of which introduces its own set of extra constraints:                
    \begin{enumerate}[label=(\alph*)]
    \item \textbf{Open \& breaking object contact} $\left(e_i \in \mathcal{S}^e_{open} \cap \mathcal{S}^e_{break} \right)$: \\
    Considering a simple mode schedule ${\bigl\{\{t\in[0,t_{open}); \ \emph{closed}\}, \ \{t \geq t_{open}; \ \emph{open}\}\bigr\}}$ that captures a contact-breaking maneuver, we introduce the following time-based constraint that is active for $t \in [t_{open}, t_{open} + \Delta_{retract}]$ with $\Delta_{retract} > 0$
        \begin{equation}
        _{\mathcal{C}_i}\hat{\bm n}^T \ \Bigl( \ _{\mathcal{C}_i}\bm \nu_{e_i}\bigl(\bm q_r\bigr) \ - \ _{\mathcal{C}_i}\bm \nu_{e_i}\bigl(\bm q_o, \bm r_{e_i}(\bm q_r)\bigr)\Bigr) \ - \ \nu_{retract}(t) = 0
        \end{equation} 
        where $\nu_{retract}(t)$ is a time-parameterized trajectory representing the reference separation velocity (e.g., trapezoidal velocity profile). 
    \item \textbf{Open \& establishing object contact} $\left(e_i \in \mathcal{S}^e_{open} \cap \mathcal{S}^e_{establish} \right)$: \\
    Considering a simple mode schedule ${\bigl\{\{t\in[0,t_{close}); \ \emph{open}\}, \ \{t \geq t_{close}; \ \emph{closed}\}\bigr\}}$ that captures a contact-making maneuver, we introduce the following time-based constraint that is active for $t \in [t_{close} - \Delta_{approach}, t_{close}]$ with $\Delta_{approach} > 0$
        \begin{equation}
        _{\mathcal{C}_i}\hat{\bm n}^T \ \Bigl( \ _{\mathcal{C}_i}\bm \nu_{e_i}\bigl(\bm q_r\bigr) \ - \ _{\mathcal{C}_i}\bm \nu_{e_i}\bigl(\bm q_o, \bm r_{e_i}(\bm q_r)\bigr)\Bigr) \ - \ \nu_{approach}(t) = 0
        \end{equation}   
        where $\nu_{approach}(t)$ is a time-parameterized reference for the approach velocity. Additional constraints are introduced here that depend on the type of object contact (i.e., nonprehensile point or surface, or prehensile point):  
    \item[i.] \textbf{Open \& establishing nonprehensile object contact} $\left(e_i \in \mathcal{S}^e_{open} \cap \mathcal{S}^e_{establish} \land c_i \in \mathcal{S}^c_{nonpreh}\right)$: \\
    For $t\geq t_{close}$ we impose the following constraints
    
    $\text{If $c_i \in \mathcal{S}^c_{point}$:} \quad $
    
        \begin{equation}
        \left\{
            \begin{split}
            &\bm r_{c_i} - \bm r_{e_i} = \bm 0_{3\times1} \quad & \textit{(matching positions)}
            \end{split}
            \right.
            \end{equation}
    $\text{If $c_i \in \mathcal{S}^c_{surface}$:} \textit{ (end-effector lies inside surface)} $            
            \begin{equation}
        \left\{
            \begin{split}
            & _{\mathcal{C}_i}\hat{\bm T}^T \bigl( \ _{\mathcal{C}_i}\bm r_{e_i} \ - \  _{\mathcal{C}_i} \overbar{\bm r}_{c_i}\bigr) - \bm b_{lower} \geq \bm 0_{2\times1} \quad &  \\
           & \bm b_{upper} \ - \ _{\mathcal{C}_i}\hat{\bm T}^T \bigl( \ _{\mathcal{C}_i}\bm r_{e_i} \ - \  _{\mathcal{C}_i} \overbar{\bm r}_{c_i}\bigr) \geq \bm 0_{2\times1} \quad & \\
             & _{\mathcal{C}_i}\hat{\bm n}^T \bigl( \ _{\mathcal{C}_i}\bm r_{e_i} - \  _{\mathcal{C}_i}\overbar{\bm r}_{c_i}\bigr) = 0 \quad &
            \end{split}
        \right.
        \end{equation}  
        where $\bm b_{lower}$ and $\bm b_{upper} \in \mathbb{R}^2$ denote the lower and upper bounds corresponding to the surface contact, and $\overbar{\bm r}_{c_i}\in \mathbb{R}^3$ is an arbitrary point belonging to the surface. To ensure the solver converges to reliable behaviors, the bounds were slightly tightened to push solutions away from the actual surface edges.      
\item[ii.] \textbf{Open \& establishing prehensile object contact} $\left(e_i \in \mathcal{S}^e_{open} \cap \mathcal{S}^e_{establish} \land c_i \in \mathcal{S}^c_{prehensile}\right)$: \\
    For $t\geq t_{close}$ we impose the following constraints
    \begin{equation}
    \left\{
        \begin{split}
        & \bm r_{c_i} - \bm r_{e_i} = \bm 0_{3\times1}  &&  \textit{(matching positions)} \\
        & _{\mathcal{C}_i}\hat{\bm n}^T \ _{\mathcal{E}_i}\hat{\bm n} - 1 = 0  &&  \textit{(matching orientations)} \\
        & _{\mathcal{C}_i}\hat{\bm b}^T \ _{\mathcal{E}_i}\hat{\bm t} = 0  && 
        \end{split}
    \right.
    \end{equation}      
    
    \end{enumerate}    
\item \textbf{Grounded contact} $\left(e_i \in \mathcal{S}^e_{ground}\right)$: \\
    We define the constraints under the assumption that all grounded contacts are nonprehensile interactions and that the terrain is flat 
    \begin{equation}
    \left\{
        \begin{split}
        & \bm \nu_{e_i} = \bm 0_{3\times1}  &&  \textit{(zero linear velocity)} \\
        & \mu_s f^z_{e_i} - \sqrt{f^{x^2}_{e_i} + f^{y^2}_{e_i}} \geq 0  &&  \textit{(force lies inside friction cone)}
        \end{split}
    \right.
    \end{equation} 
\item \textbf{Swinging foot contact} $\left(e_i \in \mathcal{S}^e_{open} \cap \mathcal{S}^e_{foot} \cap \overbar{\mathcal{S}}^e_{switch}  \right)$: 
    \begin{equation}
        \nu^z_{e_i} - \nu_{swing}(t) = 0
    \end{equation}         
    where $\nu_{swing}(t)$ is a time-parameterized trajectory representing the reference swinging velocity (e.g., cubic spline for position profile). 
\end{enumerate}


\subsection*{Section S4. Contact-driven OCP formulation: Cost Function}
The cost function of the inner-level OCP is formed of simple quadratic terms: 
\begin{equation}
    \left
    \{
    \begin{split}
        &\mathcal{L}\left(\bm x, \bm u, \Tilde{\bm r}\right) = \underbrace{\left\lVert{\bm S^T_{_1} \bm x - \overbar{\bm d}}\right\rVert^{^2}_{_{\overbar{\bm Q}}} + \left\lVert{\bm u - \overbar{\bm u}}\right\rVert^{^2}_{_{\overbar{\bm R}}}}_{regularization} + \underbrace{\left\lVert{\bm S^T_{_2} \bm x - \Tilde{\bm r}}\right\rVert^{^2}_{_{\bm Q}}}_{tracking}  &&  \textit{(intermediate cost)}\\[0.5ex]
        &\Phi\left(\bm x, \Tilde{\bm r}\right) = \underbrace{\left\lVert{\bm S^T_{_1} \bm x - \overbar{\bm d}}\right\rVert^{^2}_{_{\overbar{\bm P}}}}_{regularization} + \underbrace{\left\lVert{\bm S^T_{_2} \bm x - \Tilde{\bm r}}\right\rVert^{^2}_{_{\bm P}}}_{tracking}  &&  \textit{(terminal cost)}
    \end{split}
    \right.
\end{equation}
where $\overbar{\bm d} \in \mathbb{R}^{d}\subset\mathbb{R}^{n_x}$ holds default values for a subset of the state, $\bm S_{_1} \in \mathbb{R}^{n_x\times d}$ is a selection matrix used to extract the corresponding states from $\bm x$, and $\overbar{\bm Q}, \overbar{\bm P} \in \mathbb{R}^{d\times d}$ are positive semi-definite state-weighting matrices that are task-independent. $\overbar{\bm u} \in \mathbb{R}^{n_u}$ denotes the nominal input and $\overbar{\bm R} \in \mathbb{R}^{n_u\times n_u}$ is a positive definite input-weighting matrix that is also task-agnostic. The only task-dependent weighting matrices appear in the cost terms affected by the exogenous references $\Tilde{\bm r}$, namely $\bm Q$ and $\bm P$. The selection matrix $\bm S_{_2}$ extracts the states driven by $\Tilde{\bm r}$ from $\bm x$; the range of $\bm S_{_2}$ is disjoint from that of $\bm S_{_1}$. It is important to note that all weighting matrices are diagonal, and that we set $\overbar{\bm P} = \overbar{\bm Q}$ and $\bm P = \bm Q$ to reduce the number of tuning parameters. The cost weights adopted in the inner-level OCP are presented in \cref{tables2}.  

\subsection*{Section S5. Reference generation}
The OCP cost references are defined at the level of the base 2D-pose and the object configuration: $\Tilde{\bm r} = (\Tilde{\bm r}_{b_{xy}} \ \ \Tilde{r}_{b_{yaw}} \ \ \Tilde{\bm r}_o)\in \mathbb{R}^{3+n_o}$. Each reference corresponding to the extension of a node $n_k$ depends on its initial continuous state $\bm x_{_{T_{k}}}$ and its manipulation mode given by the pair $\langle\bm s_k, \bm a_k\rangle$. More importantly, the direction of extension is determined by a random variable $\Tilde{\bm y}\in SE(2)\times\mathbb{R}^{n_o}$ sampled from either a uniform distribution $\mathcal{U}[\Tilde{\bm r}_{min}, \Tilde{\bm r}_{max}]$ or a normal distribution $\mathcal{N}(\bm \mu, \bm \Sigma)$ whose task-specific parameters are directly linked to the definition of our goal set, $\mathcal{X}_{goal}$, as follows: 
\begin{equation}
    \mathcal{X}_{goal}:=\Bigl\{\bm x \in \mathbb{R}^{n_x} \ \Bigl| \ \left\lVert{\bm S^T_{goal}(\bm S^T_{_2} \bm x - \bm \mu)}\right\rVert \leq \delta \Bigr\}
\end{equation}
where $\delta$ is a small positive threshold, $\bm S_{_2} \in \mathbb{R}^{n_x\times (3+n_o)}$ is a selection matrix that extracts the states corresponding to the reference space, while $\bm S_{goal}\in \mathbb{R}^{(3+n_o)\times n_g}$ is used to extract the elements of interest for task-attainment from a reference vector. Moreover, we construct the covariance matrix $\bm \Sigma$ such that we end up with a low variance with respect to the non-zero coordinates of $\bm S_{goal}$ and a high variance otherwise. This results in a goal-directed distribution $\mathcal{N}$ with an exploratory aspect primarily in goal-independent directions. The related hyperparameters for different loco-manipulation scenarios are shown in \cref{tables3}.       

Given a node $n_k(\bm s, \bm a, \bm x_{{T}})$ and a randomly sampled target $\Tilde{\bm y}$, the reference $\Tilde{\bm r}$ is generated depending on the contact mode, as shown below: 
\begin{itemize}
    \item [--] \textbf{General case}:
    \begin{equation}
        \left\{
        \begin{split}
            & \Tilde{\bm r}_{b_{xy}} = \overbar{\bm r}_{b_{xy}} + T \cdot \min\Bigl(\left\lVert \Tilde{\bm v}_{b_{xy}} \right\rVert, v^{max}_{b_{xy}}\Bigr) \cdot \dfrac{\Tilde{\bm v}_{b_{xy}}}{\left\lVert \Tilde{\bm v}_{b_{xy}} \right\rVert} \\[0.75ex]
            & \Tilde{r}_{b_{yaw}} = \overbar{r}_{b_{yaw}} + T \cdot \min\Bigl(\left\lVert \Tilde{v}_{b_{yaw}} \right\rVert, v^{max}_{b_{yaw}}\Bigr) \cdot \dfrac{\Tilde{v}_{b_{yaw}}}{\left\lVert \Tilde{v}_{b_{yaw}} \right\rVert}  \\[0.75ex]
            & \Tilde{r}_{o_i} = \overbar{r}_{o_i} + T \cdot \min\Bigl(\left\lVert \Tilde{v}_{o_i} \right\rVert, v^{max}_{o_i}\Bigr) \cdot \dfrac{\Tilde{v}_{o_i}}{\left\lVert \Tilde{v}_{o_i} \right\rVert} \quad \forall i = 1,\dotsc,n_o \\
        \end{split} 
        \right.
    \end{equation}
    \item [--] \textbf{Contact switch occuring $(\mathcal{S}^e_{switch} \neq \emptyset)$}: 
    \begin{equation}
         \Tilde{\bm r} = \overbar{\bm r}
    \end{equation}
    \item [--] \textbf{No closed object contact $\bigl(\mathcal{S}^e_{object} = \emptyset\bigr)$}:
    \begin{equation}
        \Tilde{\bm r}_{o} = \overbar{\bm r}_o
    \end{equation}
    \item [--] \textbf{Stance gait - no locomotion possible $\bigl(\forall t \in [0, T]: \mathcal{S}^e_{open} \cap \mathcal{S}^e_{foot} \cap \overbar{\mathcal{S}}^e_{switch} = \emptyset\bigr)$}:
    \begin{equation}
        \left\{
        \begin{split}
            & \Tilde{\bm r}_{b_{xy}} = \overbar{\bm r}_{b_{xy}} \\
            & \Tilde{r}_{b_{yaw}} = \overbar{r}_{b_{yaw}}
        \end{split}
        \right.
    \end{equation}
\end{itemize}
where ${\overbar{\bm r} = \bm S_{_2}^T \bm x_T}$ is the initial state projected onto the reference space, while ${\Tilde{\bm v} = \bigl(\Tilde{\bm y} \boxminus \overbar{\bm r}\bigr) \bigl/ T}$ is the average velocity vector with $\boxminus := (SE(2) \times \mathbb{R}^{n_o}) \times (SE(2) \times \mathbb{R}^{n_o}) \rightarrow \mathbb{R}^{3+n_o} $ acting as a difference operator. The hyperparameters $T$ and $v^{max}_i$ correspond to the OCP time horizon and the maximum velocity magnitude for different coordinates $i$, respectively.    
\subsection*{Section S6. Implementation details}
\textbf{Solving the OCP}\\
The rapid generation of solutions for the bilevel optimization is strongly coupled to the computational effort that comes with each individual node extension. Therefore, we leverage the state-of-the-art in fast structure-exploiting OCP solvers for switched systems. Specifically, we rely on the solver introduced in \cite{PerceptiveMPCRuben} which employs a direct Multiple Shooting scheme to transform the infinite dimensional optimization into a finite dimensional NLP. The resulting NLP is then solved with an efficient implementation of the Sequential Quadratic Programming (SQP) technique that is based on a custom filter line-search algorithm \cite{Nocedal} and the \emph{HPIPM} framework \cite{HPIPM} for solving the underlying QP subproblems. Each QP is formulated with respect to a reduced-order program that arises from projecting all state-input equality constraints. Moreover, state-only equality constraints and inequality constraints are both treated as soft constraints using quadratic penalties or quadratically-relaxed log-barriers \cite{FeedbackMPCRuben}; alternatively, these can also be handled with an augmented-Lagrangian approach as in \cite{AugmentedLagrangianJPS}. Consequently, the projected QP subproblem ends up being a minimization of the augmented cost's quadratic approximation, subject to the linearized dynamics. Furthermore, all of the multi-body kinematics and dynamics involved in the OCP formulation are efficiently computed using the \emph{Pinocchio} library \cite{pinocchio1,pinocchio2}. The linear-quadratic approximation of the NLP is also partially handled with analytical-derivatives provided by \emph{Pinocchio}, and in part through automatic differentiation using the \emph{CppAD} package \cite{cppad}. We note that the same framework used for solving the short-horizon OCPs during the bilevel search was also adopted in this work for the long-horizon trajectory optimization and the online MPC module. The software implementation for the Multiple Shooting algorithm is publicly available in the \emph{OCS2} C++ toolbox \cite{OCS2}.   
\\\\
\textbf{Nearest-neighbor exploration metric} \\
To begin with, we make use of the \emph{Open Motion Planning Library} (\emph{OMPL}) \cite{OMPL} to define our nearest-neighbor search (NNS) data structures. A possible underlying distance metric that can be applied to our setting which involves a search in $SE(2)\times \mathbb{R}^{n_o}$ is presented below:
\begin{equation}
\scalemath{0.95}{  \begin{split}
    d(\bm q_{b1}, \bm q_{o1}; \bm q_{b2}, \bm q_{o2}) := \ & c_1 \left\lVert \bm q_{b2_{xy}} - \bm q_{b1_{xy}} \right\rVert_{2} + c_2 \left\lVert \bm q_{o2} - \bm q_{o1} \right\rVert_{2} + \\
    & c_3 \cdot \min{\Bigl(2\pi - |\bm q_{b2_{yaw}} - \bm q_{b1_{yaw}}|, |\bm q_{b2_{yaw}} - \bm q_{b1_{yaw}}|\Bigr)}
    \end{split}}
\end{equation}
It is worth noting that this metric would be different in the case of nonholonomic systems. Given a randomly-sampled node characterized by $(\bm q_{b2} \ \ \bm q_{o2})$, finding the nearest neighbor with configuration $(\bm q_{b1} \ \ \bm q_{o1})$ in the tree based on (20) yields an expansion of tree branches that effectively covers the search space \cite{LaValle}. However, the above metric does not account for the hybrid nature of our problem (i.e., the discrete states are not considered during the exploration). This could lead to scenarios where a particular contact state with a sparse appearance in the tree has a low probability of being selected for expansion. Therefore, in order to balance exploration among different modes, we divide the fully connected tree into $n_m$ disconnected subtree structures, where $n_m$ is the number of distinct contact states already appearing in the tree at the current stage of the bilevel search. Then each sub-structure can be sampled uniformly before applying the distance metric in (20) to find the nearest node belonging to this subtree. Such a strategy ensures that each manipulation mode is equally-likely to be explored. Nonetheless, with the aim of achieving a proper balance between exploration and exploitation, rather than arbitrarily selecting subtrees, we introduce a different criterion that is based on the UCT (Upper Confidence Bounds applied to Trees) algorithm \cite{UCT}: 
\begin{equation}
    R_{subtree_i} = \underbrace{\dfrac{e^{-C^{^*}_{n_i}/\lambda}}{N_i\cdot\sum^{n_m}\limits_{i=1} e^{-C^{^*}_{n_i}/\lambda}}}_{\text{exploitation term}} \ + \ \beta\cdot \underbrace{\sqrt{\dfrac{\ln{\Bigl(\sum^{n_m}\limits_{i=1} {N_i}}\Bigr)}{N_i}}}_{\text{exploration term}}
\end{equation}
where $\beta$ and $\lambda$ are tuning parameters, $N_i$ is the total number of attempted node extensions (i.e. both successful and failed extensions) corresponding to mode $i$, and $C^{^*}_{n_i}$ is the lowest cumulative cost discovered so far among the nodes of subtree $i$. The subtree with the highest current reward $R_{subtree}$ is chosen for expansion. \\\\
\textbf{Auxiliary pruning strategies} \\
We introduce auxiliary pruning strategies that are not essential to the main algorithm but can help maintain a sparse tree and slightly speed-up the bilevel search:  
\begin{itemize}  
\itemsep0em 
    \item [--] All uniformly random extensions of internal nodes involving a contact-switching action can be discarded due to (17) in Section S5 which results in an OCP equivalent to that already solved during the goal-directed extension phase.  
    \item [--] Additional task-dependent rules can be applied to prune out unnecessary branches, such as rejecting a contact-making event with a freely-moving object or enforcing continuous robot-object contact throughout the manipulation task. Another reason behind including such rules is that the assumption of full object-state observability in the contact-free case does not really hold, especially when relying solely on proprioceptive feedback. Therefore, we can consider this fact in the planning phase to ensure that the offline solutions are transferable to the real system.     
    \item [--] After solving an OCP characterized by a contact state $\bm s_k$ and a non-switching action, we search for the nearest neighbors in the subtree corresponding to $\bm s_k$ within a radius $\delta_{prune}$ from $\bm x_{T_{k+1}}$. Afterwards, we keep the node with lowest cumulative cost $C_n$ and eliminate the others from the tree. The pruning radius is reduced whenever a new solution is found.     
\end{itemize}
We note that all hyperparameters introduced in this section are task-independent and are fairly intuitive to tune. Their values are presented in \cref{tables5}.   
\subsection*{Section S7. Ablation study over framework components}

\begin{figure}[htb!]
    \centering
    \includegraphics[width=1\linewidth, keepaspectratio]{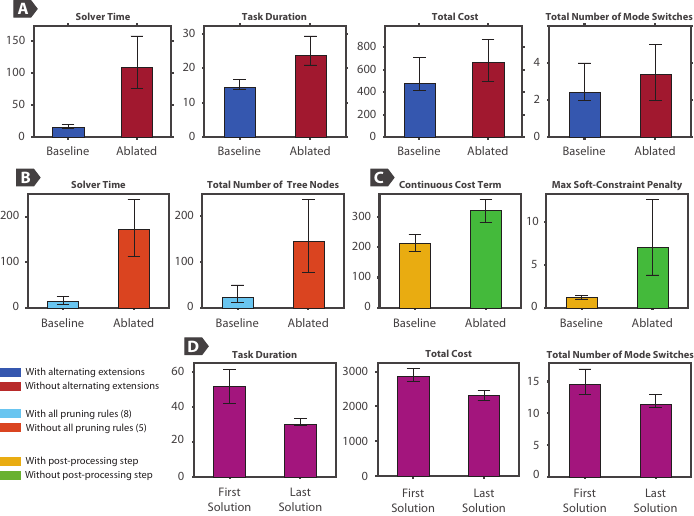}
    \caption{\textbf{Ablation study over framework components.} Each ablation study was performed using one loco-manipulation experiment \href{https://youtu.be/p25jlazh41Q}{(Movie S5)}. They include comparisons of performance metrics between our method (baseline) and an ablated version (\textbf{A}) with purely-random tree extensions, (\textbf{B}) without all pruning rules (3 are omitted), and (\textbf{C}) without a post-processing trajectory optimization step. (\textbf{D}) Comparison of performance metrics for first and last solution discovered within 4 minutes of planning time.}\label{figs3}
\end{figure}

\noindent We investigate the contributions of various key components in our framework through a series of ablation studies. All related results are presented in \cref{figs3} and Movie S5. 

First, we highlight the importance of alternating between goal-directed extensions and purely random extensions when building the tree. To this end, we run two experiments on the movable obstacle scenario wherein we sample all references either from the normal distribution $\mathcal{N}$ alone or from the uniform distribution $\mathcal{U}$. In the former, we observe that no goal-attaining solution can be found as the solver gets stuck in a bad local minimum behind a static obstacle. This is avoided with our method by allowing the algorithm to occasionally prioritize exploration. On the other hand, pure random sampling\footnote{As typically done in RRT-based planners, we also sample the goal with some small probability} does eventually yield a feasible solution; however, as shown in \cref{figs3}A, the solver is approximately 8 times slower than the baseline and results in lower-quality solutions with higher costs, longer task durations, and redundant mode switches.

Next, we evaluate the effects of incorporating logic-based interaction rules into our planner. We apply the algorithm to the dishwasher-closing example while discarding 3 out of the 8 introduced rules: Rules 1, 3, and 4. \cref{figs3}B demonstrates that although convergence is attained, the solver can be 13 times slower due to the significant increase in the number of expanded tree nodes. Moreover, as shown in Movie S5, the solutions could result in unintuitive and unnatural behaviors, such as new contacts being established but then immediately broken (i.e., due to the removal of Rule 1).

As previously mentioned in the Results section, all behaviors presented so far correspond to the first goal-attaining sequence discovered by our planner. This allowed us to ensure that even the worst (highest-cost) feasible solution was physically consistent. Here, we demonstrate the effects of applying the $\emph{ANA}^*$ strategy in our combinatorial search beyond the initially-discovered solution: We only terminate the algorithm after 4 minutes have passed. Specifically, by comparing the first and last solution for the valve-manipulation task (in the case of a high valve with 5 prehensile contacts), we observe in \cref{figs3}D that the quality of the multi-modal trajectory is ultimately enhanced. It is important to note that, in contrast to the post-processing step, which only adapts the continuous components of the solution for a fixed mode schedule, the $\emph{ANA}^*$-based search also enables optimal adjustments in the discrete sequence. These results highlight the adopted strategy's effectiveness in rapidly generating feasible sub-optimal plans (see \cref{table1}) while continuously optimizing them as the algorithm progresses.  

Furthermore, \cref{figs3}C summarizes the effects of introducing a post-processing trajectory optimization step (studied using the pull-door traversal scenario): the reduction in the overall cost and soft-constraint violation of the initial solution (i.e., initial guess) is evident. This effectively leads to visibly smoother motions (see Movie S5) with higher physical fidelity.

Movie S5 also demonstrates the importance of including an MPC layer on top of a whole-body controller. We try to track one of the hardware-tested behaviors in free motion (i.e., without tracking manipulation forces) inside a physics-based simulation using the WBC only. We observe that the execution always fails after a few seconds as the offline reference becomes unstabilizable. This can be attributed to the unmodelled disturbances induced by the feet-ground impacts, which cannot be mitigated without newly-computed references via online replanning. 

In the last experiment, we showcase the significance of tracking manipulation-force trajectories computed by the offline planner. This corresponds to the "Force tracking at manipulation contacts" element in the WBC tasks list (see \cref{tables7}). As shown in Movie S5, motion tracking alone would not be sufficient for successful task execution. Indeed, high-gain feedback control cannot fully compensate for the object's dynamics, especially when heavy objects are involved. Therefore, tracking feedforward manipulation forces that aim to cancel these dynamics under the assumption of a reasonable nominal model is essential. This highlights the benefit of holistic loco-manipulation planning at a full dynamic level rather than a kinematic one.

\subsection*{Section S8. Towards real-world deployment}

\noindent As a potential real-world application of our current framework, we consider the case of a mobile manipulator being deployed in a certain industrial facility to perform repetitive maintenance and inspection operations. In such a confined setting, reasonable knowledge about the properties of the objects of interest can be assumed. This enables us to specify suitable nominal models and affordances for the set of distinct objects available on-site, and accordingly construct a fixed library of loco-manipulation trajectories. So far, simply running these trajectories blindly from start to end has been sufficient for the sake of this paper, as the primary purpose of the hardware results has been to demonstrate the physical consistency of the generated behaviors. However, enabling a successful deployment entails additional practical considerations that are central to the autonomy and safety underlying the overall loco-manipulation pipeline.  

First, we note that during the lab experiments, we were able to directly run the pre-computed behaviors after an accurate positioning of the robot with respect to the manipulated object, matching the initial state appearing in the plans. This can be done manually by teleoperating the robot towards a predefined pose prior to execution. Alternatively, fiducial markers or motion capture systems are typically used in lab settings to properly localize and track the object. At deployment, extra steps are required as these would not be viable options. Therefore, instead of solely tracking the loco-manipulation trajectory, we embed this action as a single decoupled module within a larger autonomous mission defined in the form of an intuitive state-machine or behavior tree. This mission would consist of independent behaviors such as pure navigation, object detection and grasp-pose estimation, visual servoing, and multi-contact loco-manipulation. In what follows, we specifically focus our discussion around the task of autonomous door-opening and traversal, but a similar strategy can be adopted for different applications as well. One example of a simple mission sequence that augments independently-designed sub-modules is showcased in \href{https://youtu.be/MN3cclFLVMY}{Movie S7}: After navigating to an area in the vicinity of a specific door, a detection module\footnote{The detection module gets its inputs from an Intel RealSense L515 camera attached at the elbow.} based on a YOLOv5x network \cite{Yolov5} is used to identify both the door and the handle to subsequently provide an estimate of the handle's 3D pose and infer the door's hinge side. The latter, combined with the robot's location on the facility's map, provides the necessary information needed to deduce the proper door model. Afterwards, the robot tracks the handle, approaches it and grasps it given a grasp pose generated based on the handle's pose estimate and the corresponding door-affordance. It then turns the handle and tries to slightly open the door in both directions to identify whether it is a push/pull-door, and finally executes the proper loco-manipulation trajectory. 

A second issue with purely tracking offline trajectories is the inability to adequately react to unmodeled effects that require adaptations in the manipulation schedule for a successful task attainment. As shown in \href{https://youtu.be/2H0op0mksQo}{Movie S6}, some disturbances can be tolerated and mitigated with the proposed MPC-WBC control architecture; however, a disturbance that leads to an unrecoverable deviation in modes (e.g., complete loss of object contact due to slipping) while the robot blindly continues executing the full behavior can be problematic and unsafe in certain cases. Designing a general reactive policy that properly deals with such scenarios is outside the scope of this paper and is left for future work. Nonetheless, we provide a small extension to our autonomous door-traversal pipeline that brings us closer to a safe real-world deployment. The main purpose here is to demonstrate that strong deviations from a reference mode do not lead to complete failures and can be addressed with a simple reactive strategy. To this end, we identify a critical phase where the effects of potential disturbances are difficult to recover from. As demonstrated in \href{https://youtu.be/MN3cclFLVMY}{Movie S7}, this occurs at the point of contact-switching, where an artificial disturbance has been introduced (i.e., through the software) and is safely handled by augmenting a fallback condition that restarts the mission whenever an estimate of the door angle\footnote{The estimate is obtained via visual-based tracking of the door plane using an Intel RealSense D435 camera integrated at the front of the ANYmal base.} significantly deviates from that in the pre-computed plan.       

\clearpage
\newpage



\clearpage
\newpage

\twocolumn[{%
    \centering
    \vspace{2cm}
    \includegraphics[width=1\linewidth, keepaspectratio]{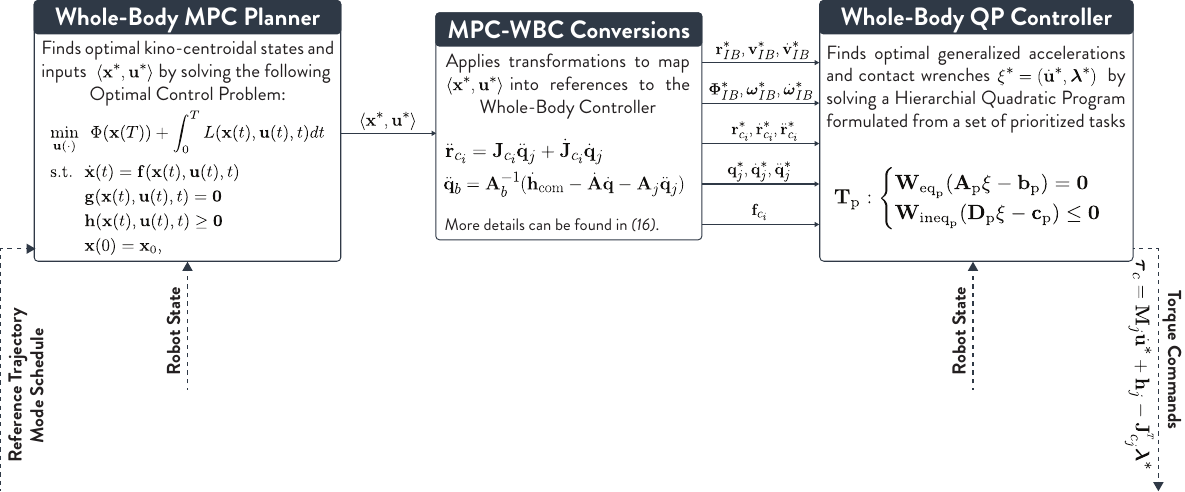}
    \captionof{figure}{\textbf{Block diagram detailing MPC-WBC interconnections.}}\label{figs4}   \vspace{2cm}
}]

\begin{table}[H]
\centering
\resizebox{0.95\columnwidth}{!}{%
\begin{tabular}{ccccc}
\toprule
\multirow{1}{*}{\bfseries Object} & \multirow{1}{*}{ \bfseries State Dimension} 
                                                                        & \multicolumn{1}{c}{\begin{tabular}[c]{@{}c@{}}\textbf{Position-Dependent}\\ \textbf{Term} $\left[\bm b_o(\bm q_o)\right]$\end{tabular}} & \multicolumn{1}{c}{\begin{tabular}[c]{@{}c@{}}\textbf{Velocity-Dependent}\\ \textbf{Term} $\left[\bm b_o(\bm v_o)\right]$\end{tabular}} & \textbf{Mass} [kg] \\ \midrule
Dishwasher      & 2                        & ---           &   $20\tanh{(20 v_o)}$                                                                                           &           25                                                                                       \\ \rule{0pt}{3ex}
Valve           & 2                       & ---           &  $\tanh{(v_o)}$                                                               &           2                                                                                    \\ \rule{0pt}{3ex}
Door            & 2                       &   $15\tanh{(15 q_o)}$    &   $10 v_o$                                                                         &     40      \\ \rule{0pt}{3ex}
Box             & 6                       & ---           &   $\begin{bmatrix}
    \sum_{i = 1}^{4} \large[\bm F_{f_i}\large]_{x,y} \\[0.6ex] \sum_{i = 1}^{4} \large[\bm r_{o,f_i} \times \bm F_{f_i}\large]_{z}
\end{bmatrix}$                                &            2       \\ 
\bottomrule
\end{tabular}}
\caption{\textbf{Objects model parameters.} The nonlinear effects in the object dynamics of (1) are split into position-dependent and velocity-dependent terms $\bigl(\text{i.e., }\bm b_o(\bm q_o, \bm v_o) = \bm b_o(\bm q_o) + \bm b_o(\bm v_o)\bigr)$. Specifically, for the box dynamic model, the box-ground frictional interaction is captured by approximating the patch contact with the four bottom vertices of the cuboid, while assuming the friction coefficient is uniform and known ($\mu = 0.7$). Accordingly, the frictional force acting at each of the four points $f_i$ is given by: $\bm F_{f_i} = 3.4\tanh{(3.4 \bm v_{f_i})}$, where $\bm v_{f_i} := \bm v_{f_i}(\bm v_o)$ is the velocity of contact $f_i$.} 
\label{tables1}
\end{table} 

\begin{table}[H]
\centering
\resizebox{0.95\columnwidth}{!}{%
\begin{tabular}{cc}

\toprule
\textbf{Elements}    & \textbf{\begin{tabular}[c]{@{}c@{}}Diagonal Weights \\ of Cost Matrices \end{tabular}} \\
\midrule
Centroidal Momentum  & {[}5, 5, 5, 5, 5, 5{]}$\ \times \ 10$                                                                  \\ \rule{0pt}{3ex}
Base 3D-Pose            & {[}10, 10, 300, 10, 40, 40{]} $\ \times \ 10$                                                             \\ \rule{0pt}{3ex}
Leg Joint Positions  & {[}2, 2, 1, 2, 2, 1, 2, 2, 1, 2, 2, 1{]} $\ \times \ 10$                                          \\ \rule{0pt}{3ex}
Arm Joint Positions  & {[}2, 2, 2, 1, 1, 1{]} $\ \times \ 10$                                                                   \\ \rule{0pt}{3ex}
Leg Contact Forces   & {[}1, 1, 1, 1, 1, 1, 1, 1, 1, 1, 1, 1{]} $\ \times \ 10^{-3}$                                                    \\ \rule{0pt}{3ex}
Arm Contact Forces   & {[}5, 5, 5{]}     $\ \times \ 10^{-3}$                                                                             \\ \rule{0pt}{3ex}
Leg Joint Velocities & {[}40, 40, 4, 40, 40, 4, 40, 40, 4, 40, 40, 4{]} $\ \times \ 10^{-1}$                      \\ \rule{0pt}{3ex}
Arm Joint Velocities & {[}40, 40, 40, 4, 4, 4{]} $\ \times \ 10^{-1}$                                                     \\ \rule{0pt}{3ex}
Dishwasher State     & {[}40, 2{]} $\ \times \ 10$                                                                                 \\ \rule{0pt}{3ex}
Valve State          & {[}20, 2{]} $\ \times \ 10$                                                                                 \\ \rule{0pt}{3ex}
Door State           & {[}40, 2{]} $\ \times \ 10$      
                   \\ \rule{0pt}{3ex}
Box State            & {[}100, 100, 1, 10, 10, 1{]}  \\                                                            
\bottomrule
\end{tabular}}
\caption{\textbf{Inner-Level OCP cost weights.} Minimal tuning effort was put into selecting these weights. Moreover, to reduce the amount of tuning parameters, we simply set the same weights for the state-dependent intermediate and terminal costs.} 
\label{tables2}
\end{table}
\clearpage
\newpage

\setlength{\tabcolsep}{10pt}
\begin{table}[H]
\vspace{3cm}
\centering
\resizebox{0.95\columnwidth}{!}{%
\begin{tabular}{ccc}
\toprule
\textbf{Loco-manipulation Scenario} & \multicolumn{2}{c}{\textbf{Goal-Directed Distribution }$\mathcal{N}$} \\ \cline{2-3} 
\multicolumn{1}{l}{}                & $\bm \mu$              & $\bm \Sigma$         \\
\midrule
Dishwasher Opening                            & (0, 0, 0, -1.5)              & diag(0.09, 0.09, 0.04, 0)                                  \\ \rule{0pt}{3ex}
Dishwasher Closing                              & (0, 0, 0, 0)              & diag(0.25, 0.25, 0.04, 0)                                   \\ \rule{0pt}{3ex}
Large Valve Turning                            & (0, 0, 0, -12.5)              & diag(0.01, 0.01, 0.01, 0)                               \\ \rule{0pt}{3ex}
Movable Obstacle Traversal                           & (2, 1, 0, 2, 1, 0)           & diag(0.04, 0.04, 0.09, 1, 1, 1)                                    \\ \rule{0pt}{3ex}
Push Door Traversal                             & (3, 0.5, 0, 1.57)              & diag(0.01, 0.01, 1, 0.09)                                   \\ \rule{0pt}{3ex}
Pull Door Traversal                               & (2, -0.35, 0, -1.2)              & diag(0.01, 0.04, 0.09, 0.64)                                   \\ \bottomrule
\end{tabular}}
\caption{\textbf{Task-dependent hyperparameters for reference generation.}}
\label{tables3}
\end{table}

\setlength{\tabcolsep}{6pt}

\begin{table}[H]
\centering
\resizebox{0.95\columnwidth}{!}{%
\begin{tabular}{cc}
\toprule
\textbf{Parameter}                     & \textbf{Value}                                                   \\
\midrule
Default gait                           & Trot / Standing-Trot (Stance during contact-making)                                                    \\ \rule{0pt}{3ex}
Inner-level OCP time horizon (T)       & \begin{tabular}[c]{@{}c@{}}Two strides (1.2 or 1.4 seconds)\end{tabular} \\ \rule{0pt}{3ex}
Maximum goal-extension iterations          & 10                                                               \\ \rule{0pt}{3ex}
Cumulative cost - Merit penalty weight  & 100                                                              \\ \rule{0pt}{3ex}
Cumulative cost - Mode switching weight & 100                                                              \\ \rule{0pt}{3ex}
Initial heuristic weight ($\alpha$)    & 10                                                               \\ \rule{0pt}{3ex}
Decay factor for $\alpha$                   & 0.8                     \\
\bottomrule
\end{tabular}}
\caption{\textbf{Task-independent hyperparameters for bilevel search.}}
\label{tables4}
\end{table}

\begin{table}[H]
\centering

\begin{tabular}{cc}
\toprule
\textbf{Parameter}        & \textbf{Value} \\
\midrule
$c_{1}$                   & 1.0            \\
$c_{2}$                   & 0.1            \\
$c_{3}$                   & 1.0            \\
$\lambda$                 & 1000           \\
$\beta$                   & 0.2             \\ 
$\delta_{pruning}$        & 0.2             \\  

\bottomrule
\end{tabular}
\caption{\textbf{Task-independent hyperparameters for NNS metric and auxiliary strategies.}}
\label{tables5}
\end{table}
 \vspace*{5\baselineskip}
\begin{table}[H]
\vspace{4cm}
\centering
\resizebox{0.95\columnwidth}{!}{%
\begin{tabular}{cc}
\toprule
\textbf{Elements}    & \textbf{\begin{tabular}[c]{@{}c@{}}Diagonal Weights \\ of Cost Matrices \end{tabular}} \\
\midrule
Centroidal Momentum  & {[}30, 30, 30, 1, 10, 10{]}                                                                 \\ \rule{0pt}{3ex}
Base 3D-Pose            & {[}80, 80, 300, 80, 80, 40{]} $\ \times \ 10$                                                             \\ \rule{0pt}{3ex}
Leg Joint Positions  & {[}2, 2, 1, 2, 2, 1, 2, 2, 1, 2, 2, 1{]} $\ \times \ 10$                                          \\ \rule{0pt}{3ex}
Arm Joint Positions  & {[}2, 2, 2, 1, 1, 1{]} $\ \times \ 10$                                                                   \\ \rule{0pt}{3ex}
Leg Contact Forces   & {[}1, 1, 1, 1, 1, 1, 1, 1, 1, 1, 1, 1{]} $\ \times \ 10^{-3}$                                                    \\ \rule{0pt}{3ex}
Arm Contact Forces   & {[}1, 1, 1{]}     $\ \times \ 10^{-3}$                                                                             \\ \rule{0pt}{3ex}
Leg Joint Velocities & {[}40, 40, 4, 40, 40, 4, 40, 40, 4, 40, 40, 4{]} $\ \times \ 10^{-1}$                      \\ \rule{0pt}{3ex}
Arm Joint Velocities & {[}40, 40, 40, 4, 4, 4{]} $\ \times \ 10^{-1}$   \\
\bottomrule
\end{tabular}}
\caption{\textbf{MPC cost weights.}}
\label{tables6}
\end{table}

\begin{table}[H]
\centering
\begin{tabular}{p{0.4375\linewidth}p{0.125\linewidth}p{0.15\linewidth}}
\toprule
\hfil Task &  \hfil Priority & \hfil Type \\
\midrule
Floating-Base dynamic EoM & \hfil 1 &\hfil Equality \\
No motion at grounded feet & \hfil 1 &\hfil Equality \\
Joint torque limits &  \hfil 1 &\hfil Inequality \\
Friction pyramid at grounded feet & \hfil 1 & \hfil Inequality \\
\noalign{\vskip 2mm} \hline\noalign{\vskip 2mm}  
Force tracking at manipulation contacts &\hfil  2 &\hfil Equality \\
Base motion tracking & \hfil 2 &\hfil  Equality \\
Swing feet trajectory tracking & \hfil 2 & \hfil Equality \\
Arm motion tracking & \hfil 2 & \hfil Equality\\
\bottomrule
\end{tabular}
\caption{\textbf{WBC prioritized tasks list.}}
\label{tables7}
\end{table}

\clearpage

\end{document}